\documentclass{IEEEtran}
\usepackage{amsmath,graphicx}
\usepackage{fancyhdr}
\usepackage{cite}
\usepackage{amsfonts}
\usepackage{amssymb}
\usepackage{verbatimbox}
\usepackage{float}

\usepackage{hyperref}
\usepackage{subfigure}
\usepackage{textcomp}
\usepackage{stfloats}
\usepackage{url}
\usepackage{verbatim}
\usepackage[commandnameprefix=always]{changes}
\usepackage{color}

\ifCLASSINFOpdf
\else
\fi

\begin{document}

\title{Unsupervised Domain Adaptation for RF-based Gesture Recognition}

\author{Bin-Bin Zhang, Dongheng Zhang, Yadong Li, Yang Hu, and Yan Chen,~\IEEEmembership{Senior~Member,~IEEE}
\IEEEcompsocitemizethanks{
\IEEEcompsocthanksitem This work was supported by National Natural Science Foundation of China under Grant 62201542, National Key Research and Development Program under Grant 2022YFC0869800, Key Research and Development Program of Anhui under Grant 2022h11020026, fellowship of China Postdoctoral Science Foundation under grant 2022M723069 and the Fundamental Research Funds for the Central Universities. (\textit{Corresponding author: Yan Chen.)}
\IEEEcompsocthanksitem B. Zhang, D. Zhang, Y. Li, Y. Chen are with the School of Cyber Science and Technology, University of Science and Technology of China, Hefei 230026, China.\ \ 
(E-mail:$\left \{\textrm{robin18, yadongli}\right \}$@mail.ustc.edu.cn, $\left \{\textrm{dongheng, eecyan}\right \}$@ustc.edu.cn).
\IEEEcompsocthanksitem Y. Hu is with the School of Information Science and Technology, University of Science and Technology of China, Hefei 230026, China.\\
(E-mail: eeyhu@ustc.edu.cn).}
}
\maketitle
\thispagestyle{fancy}

\cfoot{}
\renewcommand{\headrulewidth}{0mm}

\begin{abstract}
Human gesture recognition with Radio Frequency (RF) signals has attained acclaim due to the omnipresence, privacy protection, and broad coverage nature of RF signals. These gesture recognition systems rely on neural networks trained with a large number of labeled data. However, the recognition model trained with data under certain conditions would suffer from significant performance degradation when applied in practical deployment, which limits the application of gesture recognition systems. In this paper, we propose an unsupervised domain adaptation framework for RF-based gesture recognition aiming to enhance the performance of the recognition model in new conditions by making effective use of the unlabeled data from new conditions.
We first propose pseudo-labeling and consistency regularization to utilize unlabeled data for model training and eliminate the feature discrepancies in different domains.
Then we propose a confidence constraint loss to enhance the effectiveness of pseudo-labeling,
and design two corresponding data augmentation methods based on the characteristic of the RF
signals to strengthen the performance of the consistency regularization, which can make the
framework more effective and robust.
Furthermore, we propose a cross-match loss to integrate the pseudo-labeling and consistency regularization, which makes the whole framework simple yet effective.
Extensive experiments demonstrate that the proposed framework could achieve 4.35\% and 2.25\% accuracy improvement comparing with the state-of-the-art methods on public WiFi dataset and millimeter wave (mmWave) radar dataset, respectively.
\end{abstract}


\section{Introduction}\label{sec:introduction}
\IEEEPARstart{H}{uman} gesture recognition plays an important role in human-computer interaction systems, which could be applied to smart home, smart driving and virtual reality, etc.
Traditional approaches adopt cameras \cite{gkioxari2018detecting,wang2017recurrent}, wearable devices and phones\cite{bulling2014tutorial,guan2017ensembles}
as the sensing front-end, which however suffer from inherent drawbacks including privacy leakage, inconvenience, and limited sensing range. To address this issue, significant efforts \cite{huang2016indoor,wang2014eyes,shi2014monitoring,wang2017device,ma2018signfi,he2020wifi,zhang2019breathtrack,zhang2018multitarget,he2020non,lien2016soli,zhu2018indoor,fan2016wireless,li2020wihf,zou2016grfid,abdelnasser2018ubiquitous} have recently been made to achieve RF-based human gesture without letting the monitored subject carry any dedicated device. These systems recognize gestures by perceiving RF signals (e.g., WiFi or mmWave signals) variations caused by human gestures. 
However, the signals arriving at the receiver are not only determined by human gestures, but also significantly affected by the conditions of data collection.
As a result, there exists huge discrepancies among data collected under one condition and another, which makes the recognition system suffer from severe performance degradation when deployed under new conditions. 
\par
Generally, RF signals are affected by three factors irrelevant to gesture: the environment for data collection, the subject which performs gesture, the location and orientation of the subject, which can be summarized using the term ``domain''. 
The source domain data denotes the signals collected under specific conditions, which are adopted to train the recognition model. 
The signals collected under new conditions are referred as target domain data, which are obtained in practical deployment and unseen for the recognition model. With the data discrepancies between these two domains, the performance of existing human gesture systems are limited in real-world deployment. 

\par
To resolve this problem, 
existing methods have investigated new mathematical models to eliminate the effects of domain factors \cite{virmani2017position,zheng2019zero}, neural network architectures to extract domain-invariant features \cite{zhang2018crosssense, jiang2018towards,hayashi2021radarnet}.
However, the performance enhancements by these investigations are still limited, which makes the problem still unresolved. To tackle this challenge, we have noted that all these methods optimize the system using the labeled source domain data in a supervised manner. Due to the lack of labels in target domain, the information in the target domain data are actually wasted.
\par
In this paper, we propose an unsupervised domain adaptation framework for RF-based human gesture recognition to improve the performance on unlabeled target domain. The basic idea is to employ the pseudo-labeling and consistency regularization to make effective use of the unlabeled target domain data in an unsupervised manner. The pseudo-labeling can generate pseudo labels from unlabeled target domain data to train model, and the consistency regularization is to eliminate the feature discrepancies in different domains for enhancing the robustness of neural network. However, deploying the pseudo-labeling and consistency regularization is non-trivial and we face three challenges: \par
i) Obtaining the pseudo labels of the unlabeled target domain data and utilizing them to training the model is an important process. However, the existing works \cite{zou2019confidence,arazo2020pseudo} show that incorrect pseudo labels are inevitable due to the limitation of the recognition model and the gap between source domain and target domain, which leads to significant performance degradation of the model.\par
ii) Data augmentation methods are necessary to consistency regularization. However, unlike optical images, the semantic information in RF signals are usually not interpretable intuitively. The data augmentation methods including flipping and rotating for optical images would change the semantic information of the RF signals, which would be harmful to model training. Hence, we need to design effective data augmentation methods for RF signals.\par
iii) The proposed framework utilizes the unlabeled new domain data in an unsupervised manner. Computing and optimizing the two unsupervised losses (self-supervised loss and consistency regularization loss) would bring about the considerable computational burden, which greatly weakens the real-time performance of the system.\par
In this work, to attenuate the negative effects of incorrect pseudo labels on recognition model, we propose a confidence constraint loss to enhance the effectiveness of pseudo-labeling. To strengthen the performance of the consistency regularization, we design two corresponding data augmentation methods based on the characteristic of the RF signals. Moreover, we propose a cross-match loss integrating pseudo-labeling and consistency regularization to alleviate the computational burden.

The main contributions of our work are summarized as following:\par
(1) We propose an unsupervised domain adaptation framework for RF-based human gesture recognition, which can utilize the unlabeled target domain data to improve the performance of the recognition model.\par
(2) We propose a cross-match loss combing pseudo-labeling and consistency regularization, which could utilize the unlabeled target domain data to train in a self-supervised manner and eliminate the feature discrepancies in different domains.\par
(3) We propose a confidence constraint loss to enhance the effectiveness of pseudo-labeling, and design two corresponding data augmentation methods based on the characteristic of the RF signals to strengthen the performance of the consistency regularization, which can make the framework more effective and robust.\par
(4) We conduct comprehensive experiments on the two RF public datasets (i.e.,Wi-Fi dataset and mmWave radar dataset). The proposed framework achieves state-of-the-art results, which demonstrates that our framework is effective and universal for different kinds of RF signals.\par

The rest of this paper is organized as follows. Section \ref{related} introduces the related work. Section \ref{problem} formulates the problem. Section \ref{method} introduces the system overview, recognition model and the proposed unsupervised domain adaptation framework. Section \ref{experiments} presents the experiments on the two RF signals, the performance of the cross domain evaluations and the ablation study. Section \ref{conclusion} concludes this paper.

\section{Related Work}
\label{related}
The proposed framework is mainly related to three techniques: \emph{cross domain gesture recognition}, \emph{unsupervised domain adaptation} and \emph{semi-supervised learning}.

\subsection{Cross Domain Gesture Recognition}
Wireless human sensing techniques \cite{chen2020speednet,zhang2019calibrating,xu2017trieds,han2016enabling,zhang2020mtrack,chen2019residual} have drawn considerable attentions and made great progresses in recent years. Since human gesture recognition plays an important role in human-computer interaction, researchers have explored how to achieve gesture recognition with RF signals.
There are many prior works focusing on cross domain gesture recognition to reduce data collection and labeling efforts to generalize the recognition model, which can be roughly divided into two categories: WiFi-based and mmWave-based methods.\par
Among the WiFi-based methods, WiAG \cite{virmani2017position} and Widar3.0 \cite{zheng2019zero} construct mathematical models to derive domain-invariant features. 
CrossSense \cite{zhang2018crosssense} proposes an offline trained ANN-based roaming model mapping features from one environment to another.
EI \cite{jiang2018towards} uses an adversarial training scheme together with several constraints to generalize the model to new environments and new subjects.
For the mmWave-based models, Liu et al. \cite{liu2020real} extract dynamic variation of gestures from mmWave signals and design a lightweight CNN to recognize gestures. MmASL \cite{santhalingam2020mmasl} designs a multi-task deep neural network to achieve American sign language recognition. RadarNet \cite{hayashi2021radarnet} designs an efficient neural network and collects a large scale dataset to train a robust model. However, even though the above methods have achieved the decent performance, the progress is limited and costly. The main reason is that these methods can not make effect use of the information in the target domain data to adapt the recognition model due to lack of the labels. 
In this paper, we propose an unsupervised domain adaptation framework for RF gesture recognition to make effective use of the target domain data without labels, which can enhance the performance of the recognition model on the target domain.\par

\subsection{Unsupervised Domain Adaptation}
The Unsupervised Domain Adaptation (UDA) models assume that the source domain has sufficient labels, while the unlabeled target domain participates the model training in an unsupervised manner \cite{ben2007analysis}. 
Several approaches are adopted to learn domain-invariant features through different metrics, e.g., Maximum Mean Discrepancy (MMD) \cite{long2015learning,long2017deep}. 
Contrastive Adaptation Network (CAN) \cite{kang2019contrastive} optimizes the metric for minimizing the domain discrepancy, which explicitly models the intra-class domain discrepancy and the inter-class domain discrepancy.
Deep Adaption Network (DAN) \cite{long2015learning} adapts the high-layer features with the multi-kernel MMD criterion.
Adversarial discriminative domain adaptation (ADDA) \cite{tzeng2017adversarial} learns a discriminative representation using the source labels, and then, a separate encoding that maps the target data to the same space based on a domain-adversarial loss is used. General to Adapt (GTA) \cite{sankaranarayanan2018generate} introduces a symbiotic relation between the embedding network and the generative adversarial network.
Some other methods aim to learn a feature extractor to extract the domain-invariant features, and maintain the representation consistency through minimizing the reconstruction error between domains. For example, Domain Adversarial Neural Network (DANN) \cite{ganin2016domain} proposes a domain-adversarial training method to promote the emergence of features that are discriminative for the main learning task on the source domain. Unsupervised Image-to-Image Translation (UNIT) \cite{liu2017unsupervised} makes a shared-latent space assumption and proposes an unsupervised image-to-image translation framework based on Coupled GANs. 
Different from existing methods based on vision modality, our work focuses on how to design effective framework for RF-based gesture recognition systems. 
\par

\begin{figure*}[htbp]
	\centering
	\includegraphics[width= 1.8 \columnwidth]{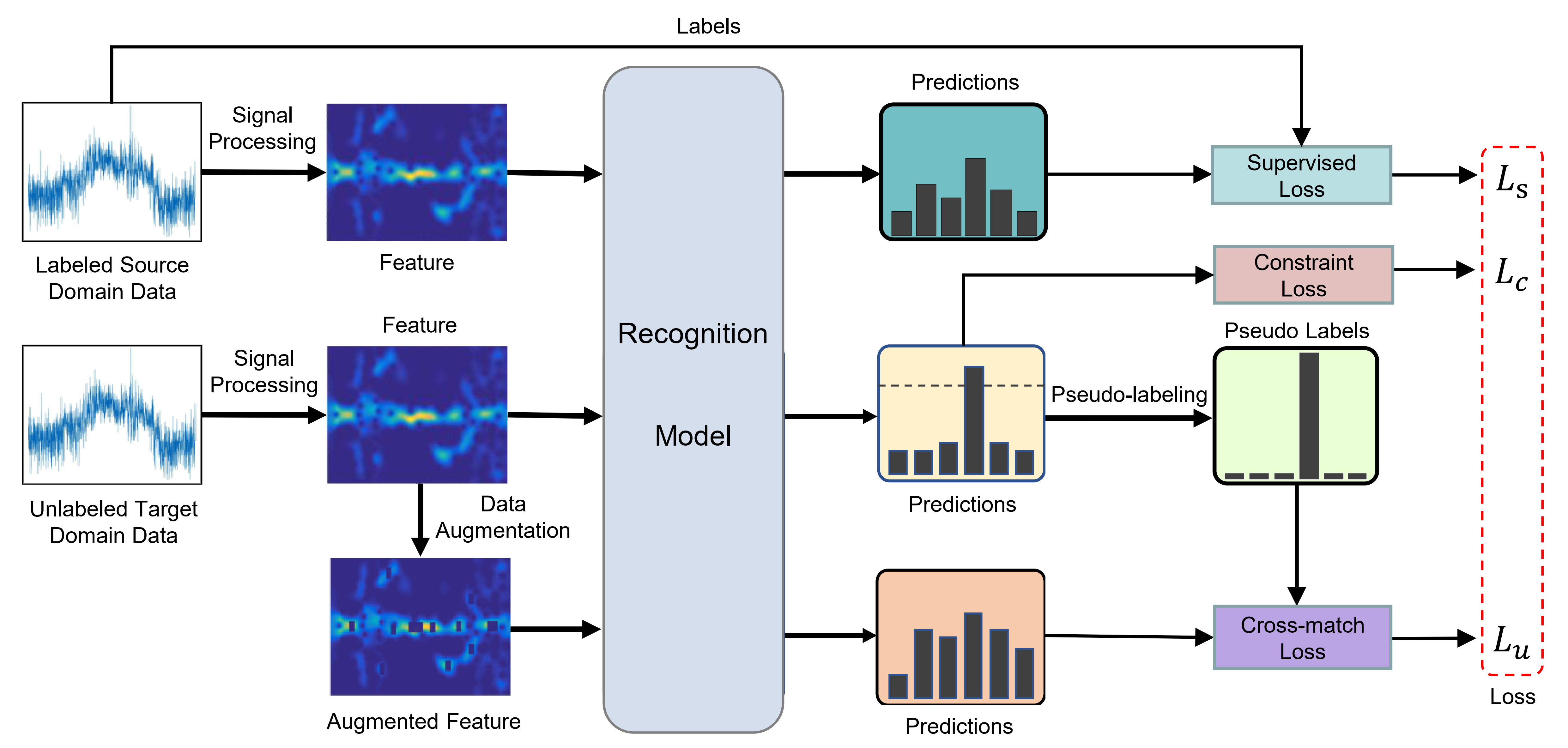}
	\caption{Overview of the framework. 
	We first augment the raw unlabeled target domain data to obtain the augmented data, then the labeled data and unlabeled data are simultaneously fed into the model to obtain three groups of predictions. After obtaining the pseudo labels from the predictions of raw unlabeled target domain data, we adopt the pseudo labels as the ground truth of the predictions of the augmented data and compute the cross-match loss $L_{u}$ between them. 
Moreover, the supervised loss $L_{s}$ between the predictions of the source domain feature and labels is computed for model training, and a confidence constraint loss $L_{c}$ is adopted to enhance the effectiveness of pseudo-labeling. During training, the three losses are optimized simultaneously.}
	\label{framework}
\end{figure*}

\subsection{Semi-Supervised Learning}
Semi-supervised learning (SSL) \cite{berthelot2019remixmatch,berthelot2019mixmatch,lee2013pseudo,sajjadi2016regularization,zhai2019s4l,sohn2020fixmatch} leverages unlabeled data to improve the performance of models when limited labeled data is provided, which alleviates the expensive labeling process efficiently.
Some recently proposed semi-supervised learning methods, such as MixMatch \cite{berthelot2019mixmatch}, FixMatch \cite{sohn2020fixmatch}, and ReMixMatch \cite{berthelot2019remixmatch} are based on augmentation viewpoints. 
MixMatch \cite{berthelot2019mixmatch} uses low-entropy labels for data-augmented unlabeled instances and mixes labeled and unlabeled data for semi-supervised learning. FixMatch \cite{sohn2020fixmatch} generates pseudo labels using the model’s predictions on weakly augmented unlabeled images. ReMixMatch \cite{berthelot2019remixmatch} uses a weakly-augmented example to generate an artificial label and enforce consistency against strongly-augmented examples.
Semi-supervised domain adaptation has more information about some target labels compared with UDA, and some related works \cite{ao2017fast,saito2019semi,qin2021contradictory,li2020online,yang2020mico} have been proposed leveraging semi-supervised signals.
Specifically, in \cite{saito2019semi}, a minimax entropy approach is proposed that adversarially optimizes an adaptive few-shot model. In \cite{qin2021contradictory}, the learning of opposite structures is unified whereby it consists of a generator and two classifiers trained with opposite forms of losses for a unified object. 
The design of the proposed framework is inspired by the principles of SSL, while we focus on domain adaption for RF signals, which has not been considered in existing works.   

\section{Problem Formulation}
\label{problem}
Given $N$ source domain gesture data $\left\lbrace \mathbf{X}_{s_{i}}, \mathbf{Y}_{s_{i}}\right\rbrace_{i=1}^{N}$ with labels and a target domain gesture data $\mathbf{X}_{t}$ without labels, the recognition model is trained using the source domain data and tested on the target domain. Especially, to enhance the performance of the recognition model on target domain, the recognition model is allowed to utilize the target domain data without the labels.
However, the target domain data can not be directly utilized for training model due to lack of labels. Thus, the recognition model trained using only the source domain data suffers from the performance degradation when applied on the target domain.
\par
To tackle the problem, an unsupervised domain adaptation framework is proposed to make effective use of the unlabeled target domain data to enhance the performance of the gesture recognition model on the target domain. 
\par

\section{Methodology}
\label{method}
Existing gesture recognition systems capture RF data frames and feed data into neural network models after pre-processing. The proposed framework is built upon this pipeline which utilizes both labeled source domain data and unlabeled target domain data to optimize the neural network model as shown in Fig. \ref{framework}. 
Specifically, we first generate pseudo labels for unlabeled target domain data with confidence control to utilize these data for model training.
We then utilize consistency regularization to eliminate the feature discrepancies in different domains. 
Furthermore, we propose a cross-match loss combing the pseudo-labeling and consistency regularization, which makes the whole framework simple yet effective. Finally, we design a confidence constraint loss and two data augmentation methods based on the characteristic of RF signals to enhancing the performance of our framework.
In the following, we will introduce our framework in detail. 
\par

\subsection{Gesture Recognition Model}
\label{recog}

\begin{figure}
	\centering
	\includegraphics[width= 1 \columnwidth]{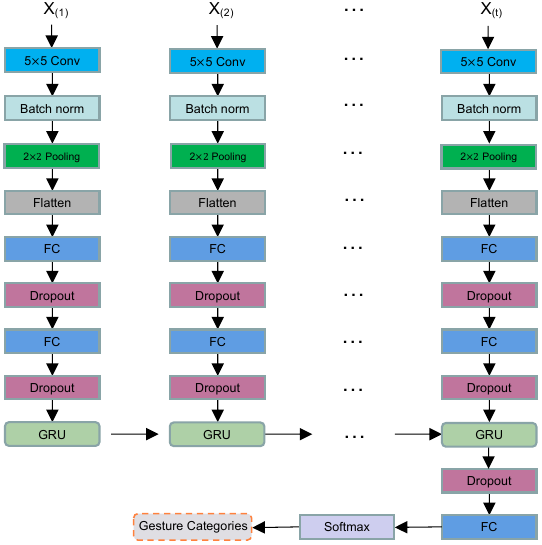}
	\caption{Structure of gesture recognition model.}
	\label{Recognition}
\end{figure}

We adopt a neural network model combing Convolutional Neural Network (CNN) layers and Gated Recurrent Units (GRU) as the gesture recognition model as shown in Fig. \ref{Recognition}
Specifically, the input data $\mathbf{X}$ are first fed into 2D CNN layers, max-pooling layers and dropout layers for feature extraction. Then, the features are flattened and fed into fully-connected layers with ReLu activation to obtain higher level representation $\left \{\mathbf{U}_{i}\right \}_{i=1}^{T}$. Finally, $\left \{\mathbf{U}_{i}\right \}_{i=1}^{T}$ is fed into GRUs to furhter extract temporal information. The output of the neural network can be expressed as 
\begin{equation}
	\mathbf{\hat{Y}} = \mathbf{G}(\mathbf{X}; \mathbf{\theta}),
\end{equation}
where $\mathbf{\theta}$ denotes the parameters of the recognition model $\mathbf{G}$, $\mathbf{\hat{Y}}$ denotes the output label of gesture category.
\par
In conventional gesture recognition systems, researchers utilize the labeled source domain gesture data $\mathbf{X}_{s}$ with label $\mathbf{Y}_{s}$ to train the recognition model then deploy the model in practical systems. 
However, due to the data discrepancies among different domains, the model trained on source domain would suffer from significant performance degradation in target domain. To resolve this problem, in the paper, an unsupervised domain adaptation framework is proposed to enhance the performance of gesture recognition model by making effective use of the unlabeled target domain RF data. 
\par

\subsection{Unsupervised Domain Adaptation Framework}
\label{unsupervised}
\subsubsection{Input and Output}
The proposed framework utilizes both the labeled source domain data and unlabeled target domain data to train the recognition model. 
We first select $B$ labeled data $\left \{\mathbf{x}_{s}, \mathbf{y}_{s}\right \}$ from the source domain gesture data pool $\left\lbrace \mathbf{X}_{s}, \mathbf{Y}_{s}\right\rbrace$ and $\mu B$ unlabeled data $\mathbf{x}_{t}$ from target domain gesture data pool $\mathbf{X}_{t}$, where $B$ is the batch size of the labeled data at every iteration, $\mu$ is a hyper-parameter.
Then we perform data augmentation on $\mathbf{x}_{t}$ to obtain data $\mathbf{x}_{t}^{aug}$. The data augmentation operation is denoted by $\mathcal{A}\left( .\right)$, which would be introduced in the following. 
After that, the source domain labeled data $\mathbf{x}_{s}$, unlabeled target domain data $\mathbf{x}_{t}$, and augmented version data $\mathbf{x}^{aug}_{t}$ are concatenated as follows:
\begin{equation}
	\label{euq5}
\begin{aligned}
	\mathbf{x}_{t}^{aug}&=\mathcal{A}\left(\mathbf{x}_{t}\right),\\
	\mathbf{X} = \mathbf{x}_{s}&\oplus \mathbf{x}_{t}\oplus \mathbf{x}_{t}^{aug},
\end{aligned}
\end{equation}
where $\oplus$ denotes the concatenation operation of the matrix. After $\mathbf{X}$ being fed into the recognition model, the predictions $\mathbf{\hat{Y}}$ can be obtained, which is composed of three parts as follows:
\begin{equation}
\label{6}
 \mathbf{\hat{Y}}= \mathbf{\hat{y}}_{s}\oplus \mathbf{\hat{y}}_{t}\oplus \mathbf{\hat{y}}_{t}^{aug},
\end{equation}
 where $\mathbf{\hat{y}}_{s}$ is the predictions of $\mathbf{x}_{s}$, $\mathbf{\hat{y}}_{t}$ represents the predictions of $\mathbf{x}_{t}$, and $\mathbf{\hat{y}}_{t}^{aug}$ denotes the predictions of $\mathbf{x}_{t}^{aug}$. \par
 
 \subsubsection{Supervised Loss}
 Due to the fact that $\mathbf{\hat{y}}_{s}$ have the labels $\mathbf{y}_{s}$, we can directly compute the cross-entropy loss between $\mathbf{\hat{y}}_{s}$ and $\mathbf{y}_{s}$ as follows:\par
 \begin{equation}
	\label{equ4}
	L_{s}=-\frac{1}{B}\sum_{i=1}^{B} \mathbf{y}_{s}^{(i)}\log\left ( \mathbf{\hat{y}}_{s}^{(i)} \right ).
\end{equation}
By computing the supervised loss $L_{s}$, the source domain data $\left\lbrace \mathbf{X}_{s}, \mathbf{Y}_{s}\right\rbrace$ are utilized to train the recognition model. The supervised loss is the component of the objective function that is adopted to compute the gradients and update the parameters during the training. Moreover, due to the $\mathbf{\hat{y}}_{t}$ and $\mathbf{\hat{y}}_{t}^{aug}$ lacking of the labels, we adopt the pseudo-labeling and consistency regularization to make effective use of them in an unsupervised manner.\par
\subsubsection{Pseudo-labeling}
To make effective use of the unlabeled target domain data, we first utilize the pseudo-labeling to generate pseudo labels of target domain data. Specifically, with the output $\mathbf{\hat{y}}_{t}$, we adopt the largest prediction score which surpasses the threshold $\tau$ as the pseudo label, which can be expressed as:
\begin{equation}
	\label{eq7}
	\mathbf{y}_{t}^{p(j)}=\left\{
	\begin{aligned}
		1 &,& \ \ \mathbf{\hat{y}}_{t}^{(j)} \geqslant \tau, \\
		0 &,& \ \ \mathrm{otherwise}.
	\end{aligned}\ \ 
	j = 1,2,...,C,
	\right.
\end{equation}
where $C$ denotes the number of gesture classes.
After obtaining the pseudo labels $\mathbf{y}_{t}^{p}$, the self-supervised loss $L_{self}$ is computed as follows:
\begin{equation}
	L_{self}=-\frac{1}{P}\sum_{i=1}^{P}\mathbf{y}_{t}^{p(i)} \cdot\log \left(\mathbf{\hat{y}}_{t}^{(i)} \right),
\end{equation}
where $P$ is the number of pseudo labels, $\mathbf{\hat{y}}_{t}$ denotes the predictions of the model on target domain.
Different from traditional methods which directly optimize the loss $L_{self}$ for model training, in this paper, we propose a cross-match loss to combine the pseudo-labeling and consistency regularization in a simple way, which aims to decrease the computation cost.

\subsubsection{Consistency Regularization}
\label{Consistency}
Apart from pseudo-labeling, we also propose consistency regularization to make effective use of the unlabeled target domain data for enhancing the robustness of neural network.
The core of consistency regularization lies in the fact that the outputs of the model should be consistent when the input data is disturbed slightly if the model is well trained. 
The conventional methods randomly disturb the data $\mathbf{x}_{t}$ to obtain the different version. 
However, \cite{xie2019unsupervised} demonstrates that adopting the data augmentation methods to transform data $\mathbf{x}_{t}$ could make the consistency regularization perform better.
Since existing data augmentation methods for visual images can not be directly applied to RF signals, we propose two data augmentation methods: \textit{Local Feature Erasing} and \textit{Time Erasing} to enhance the performance of the consistency regularization. After transforming the original data $\mathbf{x}_{t}$ into a different version $\mathbf{x}^{aug}_{t}$ by using the data augmentation methods, the original data $\mathbf{x}_{t}$ and the augmentation version $\mathbf{x}^{aug}_{t}$ are fed into the recognition model to obtain the predictions $\mathbf{\hat{y}}_{t}$ and the predictions $\mathbf{\hat{y}}^{aug}_{t}$, respectively. The discrepancies between the predictions $\mathbf{\hat{y}}_{t}$ and $\mathbf{\hat{y}}_{t}^{aug}$ are measured using cross entropy, which are referred as the consistency regularization loss $L_{reg}$. The consistency regularization loss $L_{reg}$ is computed as follows:
\begin{equation}
	\label{equ10}
	L_{reg}=-\frac{1}{P} \sum_{i=1}^{P} \mathbf{\hat{y}}_{t}^{(i)} \cdot \log\left ( \mathbf{\hat{y}}_{t}^{aug(i)} \right ).
\end{equation}
Similar to the self-supervised loss $L_{self}$, we do not directly adopt the consistency regularization loss $L_{reg}$ to optimize the recognition model, but integrate them using cross-match loss, which would reduce the computation loads.

\subsubsection{Cross-match Loss}
\begin{figure}[htbp]
	\begin{center}
		\includegraphics[scale=0.85]{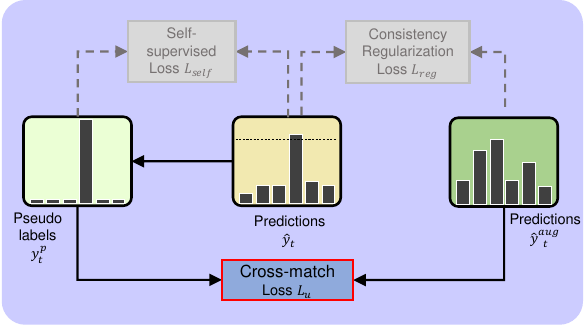}
		\caption{The cross-match loss $L_{u}$ is obtained by integrating the self-supervised loss $L_{self}$ and the consistency regularization loss $L_{reg}$.}
		\label{cross-match loss}
	\end{center}
\end{figure}

While minimizing $L_{self}$ and $L_{reg}$ during training can make effective use of the unlabeled target domain data, 
optimizing the consistency regularization loss and self-supervised loss separately would greatly increase the computation loads of the proposed framework. To resolve this problem, we have noted that 
both two losses contain the predictions $\mathbf{\hat{y}}_{t}$, which looks like a ``bridge'' connecting the pseudo labels $\mathbf{y}_{t}^{p}$ and the predictions $\mathbf{\hat{y}}^{aug}_{t}$. Inspired by this, we take out the ``bridge'' $\mathbf{\hat{y}}_{t}$ from the two losses and directly compute the discrepancies between the pseudo labels $\mathbf{y}_{t}^{p}$ and the predictions $\mathbf{\hat{y}}_{t}^{aug}$, which intuitively is equivalent to merge $L_{self}$ and $L_{reg}$ into one loss namely cross-match loss $L_{u}$ as Fig. \ref{cross-match loss} shown. The loss $L_{u}$ is computed as follows:
\begin{equation}
	\label{equ11}
	L_{u}=-\frac{1}{P} \sum_{i=1}^{P} \mathbf{y}_{t}^{p(i)} \cdot \log\left ( \mathbf{\hat{y}}_{t}^{aug(i)} \right ),
\end{equation}
where $P$ is the number of pseudo label, $\mathbf{y}_{t}^{p(i)}$ is the pseudo labels, $\mathbf{\hat{y}}_{t}$ denotes the predictions of augmentated version data. 

\subsubsection{Confidence Constraint Loss}
The confidence constraint loss $L_{c}$ is designed to attenuate the negative effects of incorrect pseudo labels on recognition model. In the proposed framework, obtaining the pseudo labels of the unlabeled target domain data and utilizing them to training the model is effective. However, the existing work shows that incorrect pseudo labels is inevitable due to the limitation of the recognition model and the gap between source domain and target domain.
According to \cite{zou2019confidence,arazo2020pseudo}, utilizing the incorrect pseudo labels to train the recognition model would suffer from significant performance degradation. 
Moreover, the early work \cite{bagherinezhad2018label} shows that enforcing a model to be very confident on only one of the class during training with the pseudo labels can hurt the learning behavior.\par
\begin{figure}[htbp]
	\begin{center}
		\includegraphics[scale=1.4]{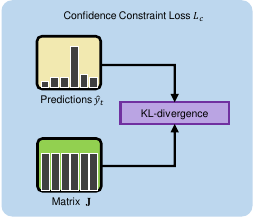}
		\caption{Confidence Constraint Loss $L_{c}$: the Kullback-Leibler (KL) divergence between the predictions $\mathbf{\hat{y}}_{t}$ and the matrix $\mathbf{J}$.}
		\label{confidence loss}
	\end{center}
\end{figure}

Thus, we design a confident constraint loss to encourage the smoothness of output probabilities and prevent overconfident prediction during training.
Specifically, as shown in Fig. \ref{confidence loss}, we compute Kullback-Leibler (KL) divergence between the predictions $\mathbf{\hat{y}}_{t}$ and the matrix $\mathbf{J}$ to measure the discrepancies between the matrix $\mathbf{J}$ and the predictions $\mathbf{\hat{y}}_{t}$, where $\mathbf{J}$ is a matrix where every entry is equal to one. Adopting the KL divergence as the confidence constraint loss and minimizing the loss during training would increase the similarity between the matrix $\mathbf{J}$ and the predictions $\mathbf{\hat{y}}_{t}$, which encourages the smoothness of $\mathbf{\hat{y}}_{t}$ like the distribution of $\mathbf{J}$. The loss is computed as follows:\par
\begin{equation}
	L_{c}=\frac{1}{P}\sum_{i=1}^{P}\mathbf{J} \cdot\log\left(\mathbf{J}\right)-\mathbf{J} \cdot\log \left(\mathbf{\hat{y}}_{t}^{(i)} \right),
\end{equation}
where $P$ is the number of pseudo labels, $\mathbf{\hat{y}}_{t}$ denotes the predictions of the model on target domain.
\subsubsection{Data Augmentation Methods}
In this paper, data augmentation methods are adopted to obtain the augmented version of the unlabeled target domain data, which is necessary to the consistency regularization. 
Unlike optical images, the semantic information in RF signals are usually not interpretable intuitively, and the data augmentation methods \cite{shorten2019survey} including flipping and rotating for optical images would change the spatial information in RF signals, which would be harmful to model training.
Moreover, we note that the features pre-processed from RF signals contain a lot of random noises and multi-path effects in spatial dimension, and plenty of redundant information in temporal dimension. Therefore, we propose the two data augmentation methods: \emph{Local Feature Erasing} and \emph{Time Erasing} to strength the effectiveness of the consistency regularization.\par
\begin{figure}[htbp]
	\centering
	\subfigure[Local feature erasing.]{
		\includegraphics[width=0.46 \columnwidth]{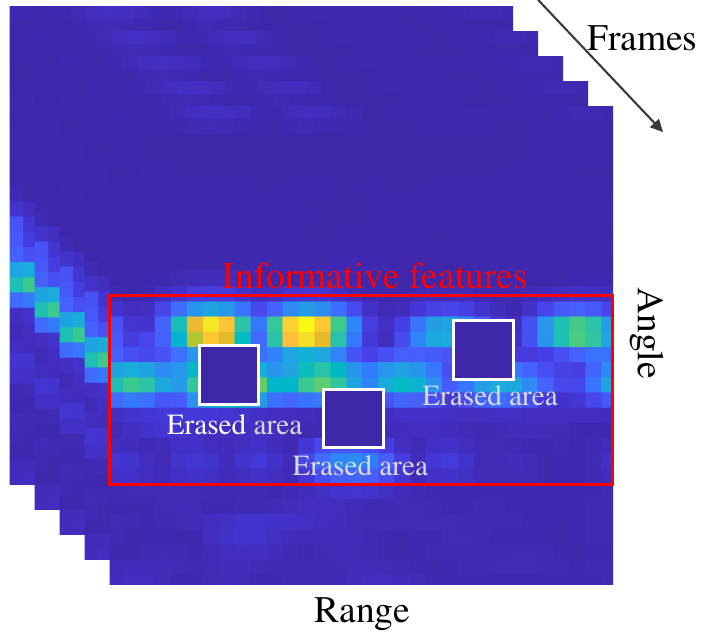}
		\label{local earasing}
	}
	\subfigure[Time erasing.]{
		\includegraphics[width=0.46 \columnwidth]{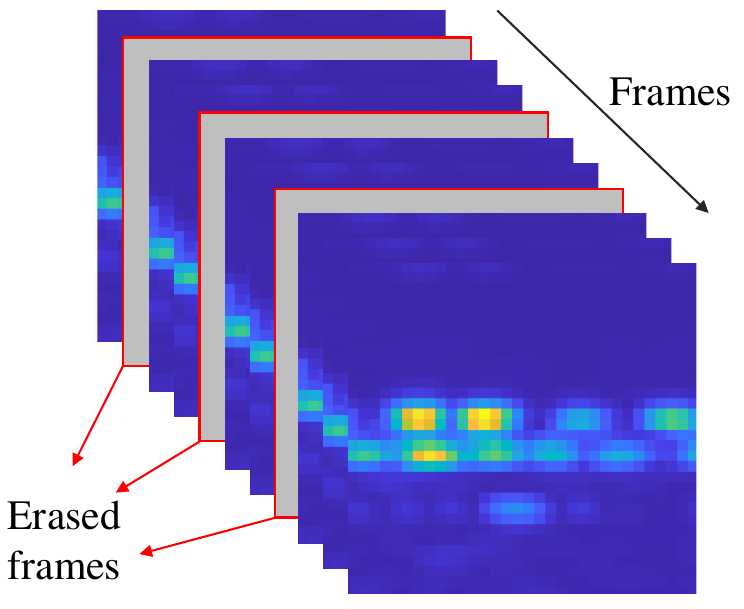}
		\label{time erasing}
	}
	\caption{Our two data augmentation methods.}
	\label{data augmentation methods}
\end{figure}

(1) Local Feature Erasing: 
During training, the neural networks tend to pay more attention to the local features containing too much domain information, which leads to the performance degradation. To address the problem, we force the recognition model to focus on the overall features associated with the gesture category by removing the local features. Specifically, In Fig. \ref{local earasing}, after the informative features (marked by red box) being detected using CFAR \cite{robey1992cfar} algorithm, part of them are selected to erase by making the value be zero, which are denoted by the erased areas.
\par

(2) Time Erasing: 
Deep neural networks are prone to overfitting without sufficient data, which usually leads to disappointing performance.
Increasing the number and diversity of gesture samples is helpful to model training. An effective method is erasing the frames in time dimension, which simulates gesture in different velocities. As Fig. \ref{time erasing} shown, the time erasing selects some frames to remove and padding with zero, which are denoted by the gray frames in Fig. \ref{time erasing}.
\par

\subsection{Objective and Training}
\label{obj}
The final objective function is composed of supervised loss $L_{s}$, cross-match loss $L_{u}$, and constraint loss $L_{c}$. The objective function is defined as follows:
\begin{equation}
	L=L_{s}+\lambda L_{u}+\eta L_{c},
\end{equation}
where $\lambda,\eta$ are predefined weight of the losses.
During training, we compute and minimize the objective function to update the parameters of neural network according to SGD \cite{qian1999momentum}.\par

\section{Experiments}
\label{experiments}
In this section, we conduct extensive experiments on two different RF-based gesture recognition datasets, i.e., WiFi and mmWave datasets, to evaluate the performance of the proposed framework. 
\subsection{Experiment with WiFi Signals}
\subsubsection{Dataset}
\begin{figure}[!ht]
	\centering
	\subfigure[Layouts of three evaluation environments.]{
		\includegraphics[width=0.26 \columnwidth]{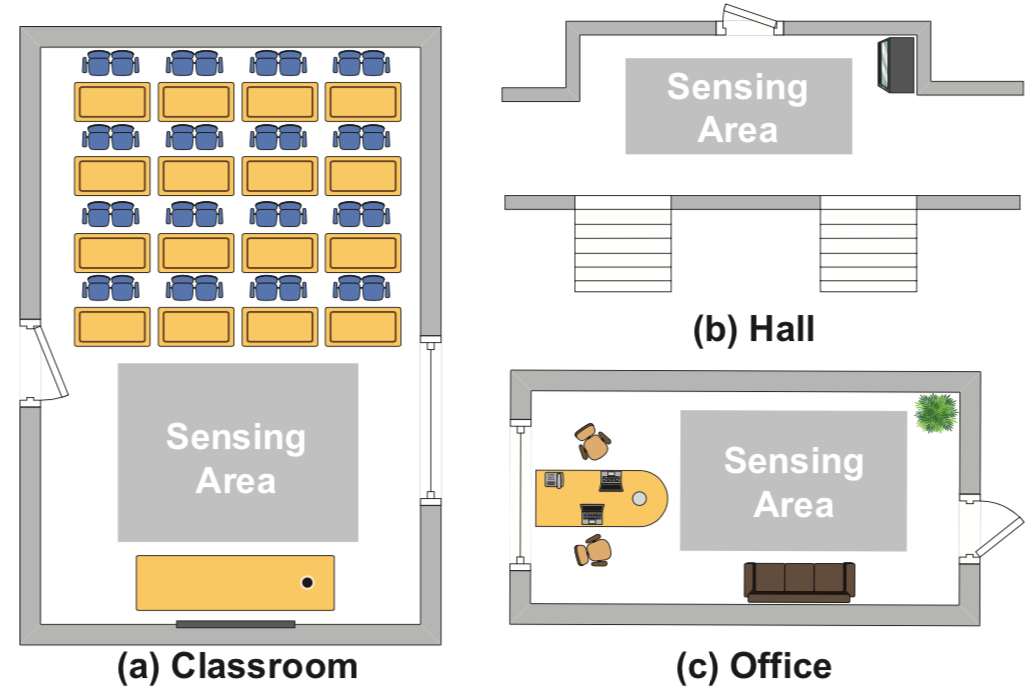}
		\label{layouts}
	}\quad
	\subfigure[A typical setup of devices and domains in one environment.] {
		\label{location and orientation}     
		\includegraphics[width=0.26 \columnwidth]{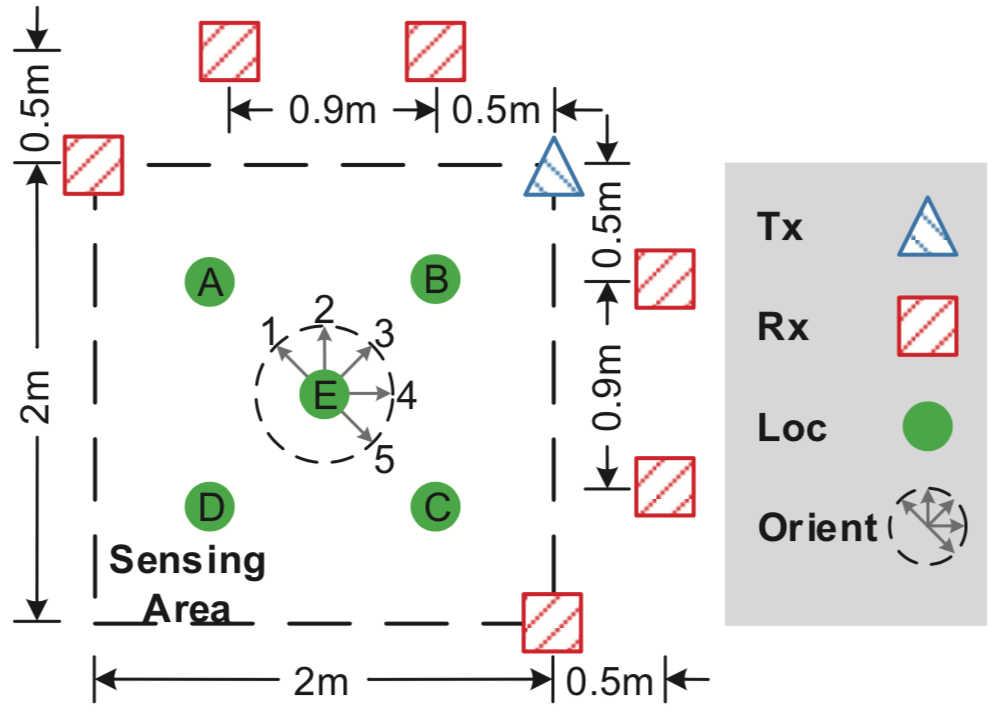}  
	}\quad
	\subfigure[Sketches of gestures evaluated in the dataset.] {
		\includegraphics[width=0.26 \columnwidth]{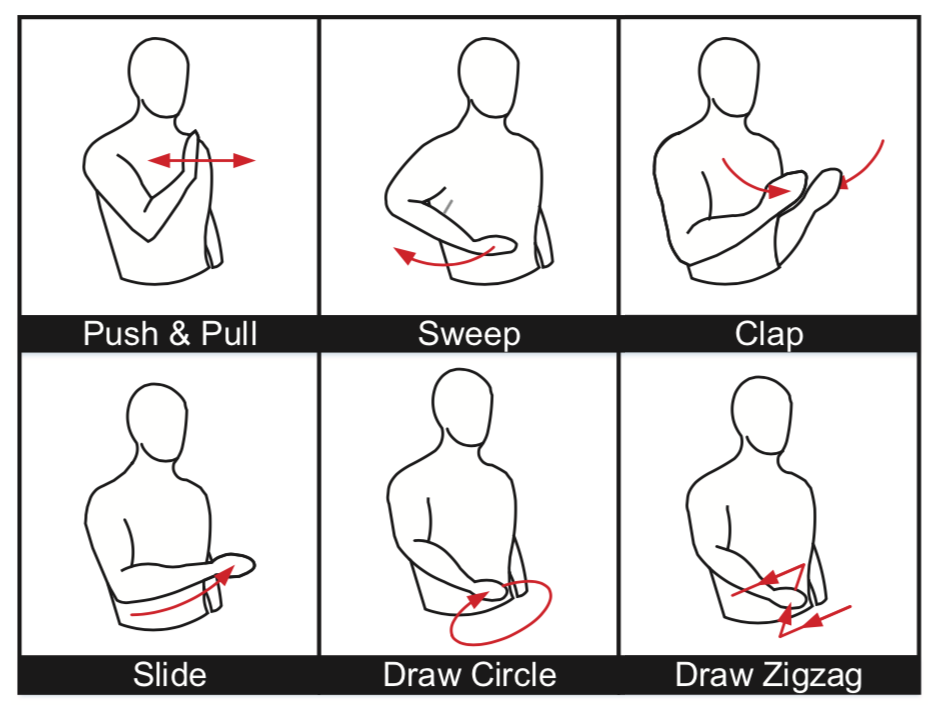}
		\label{gestures}
	}
	\caption{The setup of WiFi dataset \cite{zheng2019zero}.}
	\label{cong}
\end{figure}

We evaluate the proposed framework on the public Widar 3.0 \cite{zheng2019zero} dataset. The dataset is collected from 3 rooms, 16 users, 5 locations and 5 orientations, which contains 6 gesture categories (i.e., Push \& Push, Sweep, Clap, Slide, Draw circle and Draw zigzag) as shown in Fig. \ref{gestures}. 
In total, the dataset includes 11250 data samples (15 users $\times$ 5 locations $\times$ 5 orientations $\times$ 6 gestures $\times$ 5 instances). The layouts and sensing areas of the three different experiment environments (i.e., classroom, hall and office) are shown in Fig. \ref{layouts}, and the locations and orientations of subjects are shown in Fig \ref{location and orientation}.
\subsubsection{Data Preprocessing}
In this experiment, the WiFi signals are first processed  to the body coordinates profile (BVP) feature as in Widar 3.0 \cite{zheng2019zero}. BVP is the matrix with dimension as $V_{x} \times V_{y} \times T$, where $V_{x}$ and $V_{y}$ are the number of possible values of velocity components decomposed along each axis of the body coordinates, T is the number of BVP snapshots.\par
\begin{figure}[htbp]
	\begin{center}
		\includegraphics[scale=0.6]{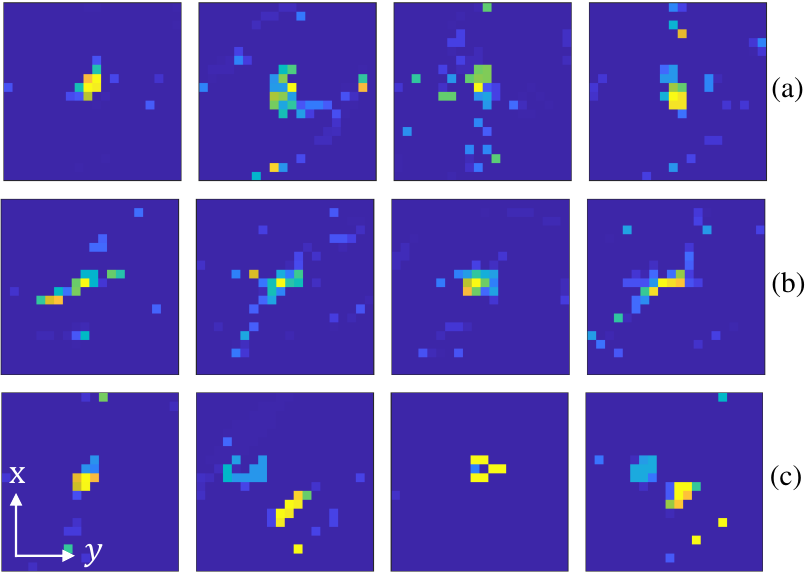}
		\caption{The BVP of the gesture draw circle and draw zigzag. (a) BVP of the gesture draw circle; (b) BVP of the gesture draw zigzag; (c) BVP of the gesture draw zigzag in other location; Columns represent time series of 4 frames. In BVP, pixel color, x-axis and y-axis correspond to doppler power, horizontal velocity, vertical velocity, respectively.}
		\label{gesture difference wifi}
	\end{center}
\end{figure}

To better understand the BVP and show the differences between the WiFi-based gestures, we randomly select the BVP of the two gestures (Draw circle and Draw zigzag) and visualize them in Fig. \ref{gesture difference wifi}. Fig. \ref{gesture difference wifi} (a) and Fig. \ref{gesture difference wifi} (b) show the differences in draw circle and draw zigzag, which indicates there are differences in spatial dimension and temporal dimension between different gestures. Moreover, Fig. \ref{gesture difference wifi} (b) Fig. \ref{gesture difference wifi} (c) denote the same gesture (draw zigzag) but collected from different locations. We easily find the discrepancies in the gesture from different domains, which are harmful to our recognition model.

\subsubsection{Baseline Methods}
We compare our approach with three deep learning models including CrossSense \cite{zhang2018crosssense}, EI \cite{jiang2018towards} and Widar3.0 \cite{zheng2019zero}. Specifically, CrossSense proposes an ANN-based roaming model to translate signal features from source domains to target domains, and employs multiple expert models for gesture recognition. 
EI is the domain adaptation scheme based on adversarial learning.
In Widar 3.0 \cite{zheng2019zero}, the authors extract the domain invariant feature BVP based on DFS data and feed them into a model combining CNN and GRUs, which is state-of-the-art method. 
 
\subsubsection{Cross Domain Evaluation}
We evaluate the performance of our framework on cases crossing different domain factors, including orientation, location, subject, and environment. When we evaluate on each domain factor, we keep the others unchanged. 
For each domain factor, we adopt one domain as the target domain, and the other domains as the source domain. We train the recognition model using source domain data with labels and target domain data without labels, then test the recognition model on the target domain. We conduct experiments using all the baseline methods for comparison.
\par
\begin{figure}[htbp]
	\begin{center}
		\includegraphics[scale=0.53]{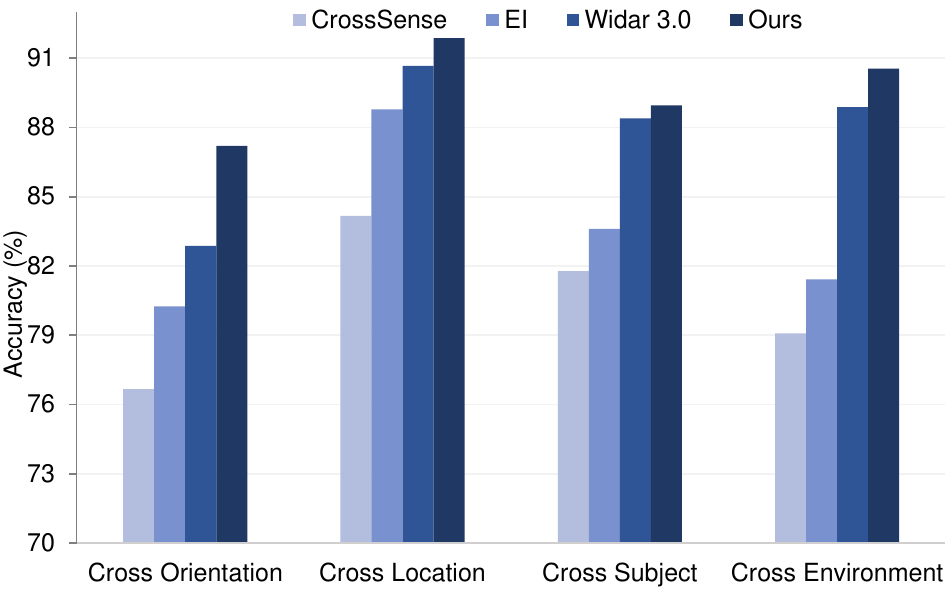}
		\caption{The overall performance of our framework on WiFi dataset.}
		\label{cross domain wifi}
	\end{center}
\end{figure}

The overall performance of our framework on WiFi dataset in four domains is shown in Fig. \ref{cross domain wifi}. In the figure, the accuracy denotes the average accuracy of every domains and we can see that our framework achieves the highest accuracy in all domains comparing with other baseline methods, which demonstrates the effectiveness of our framework on WiFi dataset. Besides, the detailed performance of our framework in four domains (orientation, location, subject, and environment) are shown below. The numbers or letters behind the arrow $\rightarrow$ represent the target domain.
\par
\emph{(1) Cross Orientation:} In this experiment, we adopt one orientation domain as the target domain and the other 4 orientations as the source domains. Table \ref{across orientation} shows that the proposed framework has an average accuracy of 87.21\%, which is the highest average accuracy comparing to other methods.
We also note that the framework has an improvement of 4.35\% over Widar 3.0 and achieves accuracy over 90\% on orientation 2 and 3.
Even though the proposed framework achieves significant performance improvements on all orientations, the accuracy of orientation 5 is not very high, the reason is that the subjects in orientation 5 are too far away from the center, which are not captured full information of gestures by the WiFi devices.  
\par
\begin{table}[htbp]
	\small
	\renewcommand{\arraystretch}{1.5}
	\caption {Cross Orientation Accuracy (\%)}
	\label{across orientation}
	\centering
	\scalebox{0.9}{
		\begin{tabular}{c|c|c|c|c|c|c}
			\hline
			Methods&$\rightarrow$ 1&$\rightarrow$ 2&$\rightarrow$ 3&$\rightarrow$ 4&$\rightarrow$ 5&Avg\\
			\hline
			CrossSense &69.41 &80.23 &80.45 &83.18 &70.07&76.67\\
			EI &70.42&81.15 &88.16 &90.83 &70.74 &80.26\\
			Widar 3.0 &80.24&89.64&89.05&82.75&72.64&82.86\\
			\hline
			Ours&\textbf{87.44}&\textbf{91.64}&\textbf{91.98}&\textbf{85.46}&\textbf{79.53}&\textbf{87.21}\\
			\hline
	\end{tabular}}
\end{table}
\emph{(2) Cross Location:} Similar to the cross orientation settings, we adopt data from 4 locations as the source domains, and the data from the last location as the target domain. 
The results are shown in Table \ref{across location}.
Compared with other methods, our framework achieves the highest average accuracy of 91.87\%. 
And location E has the highest accuracy of 95.86\%. We also note that the accuracy of all locations is above 90\%.
\par
\begin{table}[htbp]
	\small
	\renewcommand{\arraystretch}{1.5}
	\caption {Cross Location Accuracy (\%)}
	\label{across location}
	\centering
	\scalebox{0.9}{
		\begin{tabular}{c|c|c|c|c|c|c}
			\hline
			Methods&$\rightarrow$ A&$\rightarrow$ B&$\rightarrow$ C&$\rightarrow$ D&$\rightarrow$ E&Avg \\
			\hline
			CrossSense &88.61 &85.23 &82.89 &78.94 &85.16&84.16 \\
			EI &87.91&90.13 &86.85 &85.82 &93.14 &88.77\\
			Widar 3.0 &89.14&89.10&90.50&89.50&95.15&90.67\\
			\hline
			Ours&\textbf{90.47}&\textbf{90.30}&\textbf{92.30}&\textbf{90.44}&\textbf{95.86}&\textbf{91.87}\\
			\hline
	\end{tabular}}
\end{table}

\emph{(3) Cross Subject:} Data collected from different subjects may have discrepancies due to their various behavior patterns and body shapes, which weakens the performance of recognition model on different subjects.
 To evaluate the performance of the framework in subject domain, we conduct the experiments. To be specific, we adopt the data from 1 user as the target domain and 6 users' data as the source domains. The results in Table \ref{across subject wifi} show our framework achieves the highest average accuracy of 88.96\%.
 \par
\begin{table}[htbp]
	\renewcommand{\arraystretch}{2.2}
	\caption {Cross Subject Accuracy (\%)}
	\label{across subject wifi}
	\centering
	\scalebox{0.65}{
		\begin{tabular}{c|c|c|c|c|c|c|c|c|c}
			\hline
			Methods&$\rightarrow$U10&$\rightarrow$ U11&$\rightarrow$ U12&$\rightarrow$ U13&$\rightarrow$ U14&$\rightarrow$ U15&$\rightarrow$ U16&$\rightarrow$ U17&Avg\\
			\hline
			CrossSense&77.67&81.35&80.24&83.73&85.97&82.43&80.19&82.59&81.77\\
			EI&80.11&83.56&85.91&82.05&84.36&85.44&81.23&86.21&83.61\\
			Widar 3.0&86.26&\textbf{90.13}&\textbf{83.33}&89.86&87.73&89.97&\textbf{91.46}&88.34&88.38\\
			\hline
			Ours&\textbf{86.79}&89.73&80.93&\textbf{91.60}&\textbf{88.13}&\textbf{92.51}&91.33&\textbf{90.67}&\textbf{88.96}\\
			\hline
	\end{tabular}}
\end{table}
\emph{(4) Cross Environment:} To evaluate the performance of our framework in environment domain, we use the gesture data collected in 2 rooms as the source domains data and the other one as the target domain. 
The results in Table \ref{across environment wifi} show that the framework has an overall accuracy of 90.55\%, which is the highest average accuracy.\par
\begin{table}[htbp]
	\small
	\renewcommand{\arraystretch}{1.3}
	\caption {Cross Environment Accuracy (\%)}
	\label{across environment wifi}
	\centering
	\scalebox{0.9}{
		\begin{tabular}{c|c|c|c|c}
			\hline
			Methods&$\rightarrow$ Room 1&$\rightarrow$ Room 2&$\rightarrow$ Room 3&Avg\\
			\hline
			CrossSense &72.78&81.75&82.76&79.09\\
			EI &75.73&84.54&83.98&81.42\\
			Widar 3.0 &81.58&93.47&91.08&88.87\\
			\hline
			Ours&\textbf{83.93}&\textbf{94.88}&\textbf{92.85}&\textbf{90.55}\\
			\hline
	\end{tabular}}
\end{table}\par
In summary, our framework achieves state-of-the-art performance on WiFi dataset in four domains (locations, orientations, subjects and environments), which demonstrates the effectiveness of our framework on WiFi dataset. Furthermore, our framework not only achieves the considerable results on WiFi dataset, but also performs very well on mmWave dataset, which is presented below.\par 
\subsection{Experiment with mmWave Signals}
\subsubsection{Dataset}
\begin{figure}[htbp]
	\centering
	\includegraphics[width= 0.6 \columnwidth]{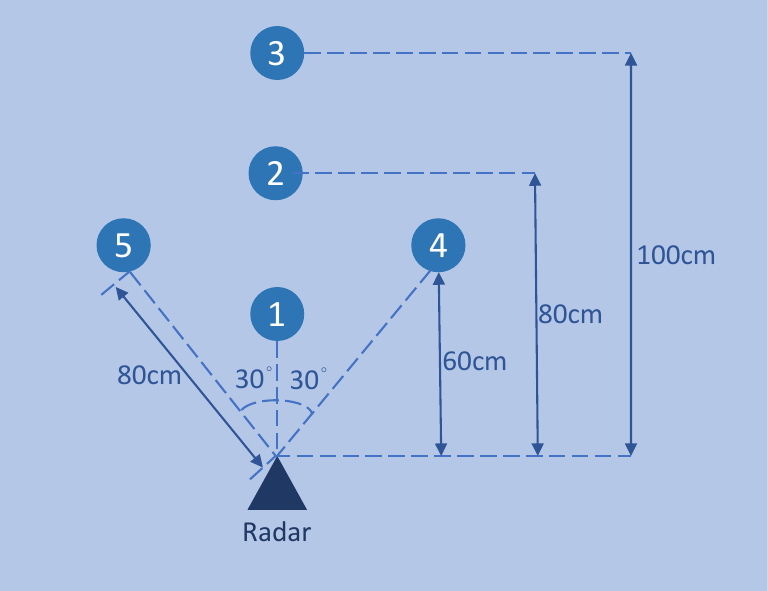}
	\caption{The setup of 5 locations in each environment \cite{li2022towards}.}
	\label{5loc}
\end{figure}

We  adopt a public mmWave dataset \cite{li2022towards}, which is collected from 25 volunteers, 6 environments and 5 locations to evaluate our framework. Fig. \ref{5loc} illustrates the setup of 5 locations in each environment. The dataset contains 6 common gestures including Push, Pull, Left swipe, Right swipe, Clockwise turning and Anticlockwise turning. Each volunteer is asked to perform each kind of gesture 5 times at each location. In total, the dataset includes 10800 data samples.
\subsubsection{Data Preprocessing}
\begin{figure}[htbp]
	\begin{center}
		\centering
		\includegraphics[scale=0.25]{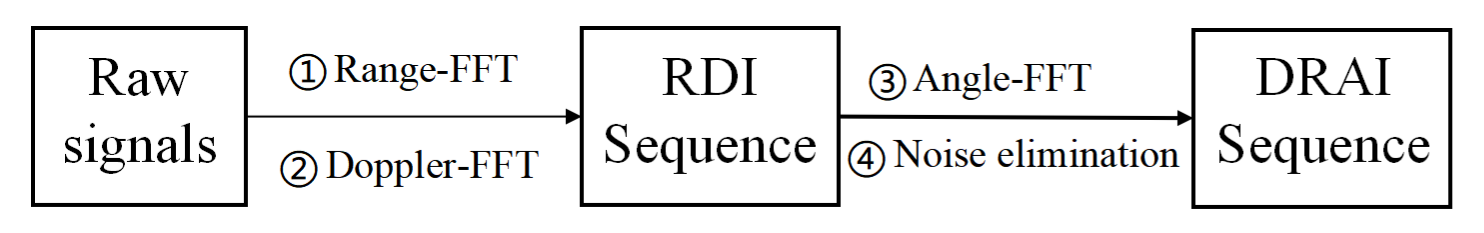}
		\caption{The calculation process of DRAI sequences.}
		\label{process}
	\end{center}
\end{figure}

In this experiment, the mmWave signals are processed to the Dynamic Range Angle Image sequence (DRAI) as in \cite{li2022towards}, which is the matrix with dimension $N\times M \times T$.
Specifically, as shown in Fig. \ref{process}, the 3D-FFT is adopted on raw signals to derive the ranges, velocities and angles of hands. Then, the noise elimination is used to filter environmental interference and improve the robustness of the recognition model. After that, we adopt the DRAI as the input of the recognition model.
\par
\begin{figure}[htbp]
	\begin{center}
		\includegraphics[scale=0.6]{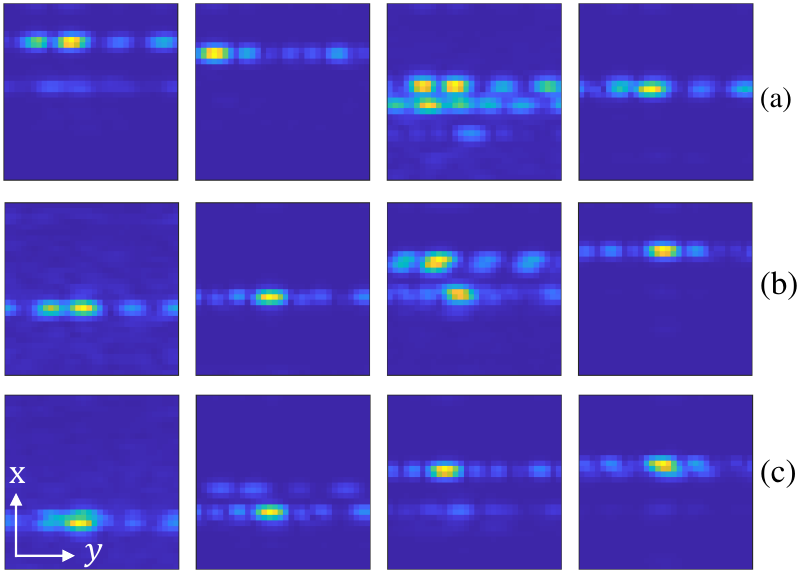}
		\caption{DRAI of the gesture slide right and slide left. (a) DRAI of the gesture slide right; (b) DRAI of the gesture slide left; (c) DRAI of the gesture slide left in other location; Columns represent time series of 4 frames. In DRAI, pixel color, x-axis and y-axis correspond to doppler power, AoA, range, respectively.}
		\label{gesture difference radar}
	\end{center}
\end{figure}

To better describe the DRAI and show the differences in different gestures and different domains, we randomly visualize the DRAI of the two gestures (e.g.,  Slide right and Slide left) in Fig. \ref{gesture difference radar}. Fig. \ref{gesture difference radar} (a) and Fig. \ref{gesture difference radar} (b) denote the two gesture collected from same settings, from which, we can see that the feature information of a gesture
consists of temporal and spatial information. Furthermore, Fig. \ref{gesture difference radar} (b) Fig. \ref{gesture difference radar} (c) represent the same gesture (slide left) collected from different locations, in which, the discrepancies are observed.
\subsubsection{Baseline Methods}
We compare our approach with two deep learning models RadarNet \cite{hayashi2021radarnet} and Widar3.0 \cite{zheng2019zero}. Specifically, RadarNet designs an efficient neural network and trains a robust model. We first process the mmWave radar signals to Range Doppler Image, then use the Range Doppler Image as the RadarNet model's input. In Widar 3.0 paper, the authors feed the features into a gesture recognition model combining CNNs and GRUs. In this paper, we adopt the gesture recognition model as the another baseline and utilize the DRAI as the input. 
\subsubsection{Cross Domain Evaluation}
We evaluate the performance of the proposed framework on mmWave dataset in different domains including location, subject, and environment. When evaluating on each domain factor, we keep the other domain factors unchanged. 
For each domain factor, we adopt one domain as the target domain and the others as the source domain.
We utilize the source domain data with labels and the target domain data without labels to train the recognition model, and test the recognition model on the target domain. We conduct experiments using all the baseline methods to validate the effectiveness of our framework.\par
\begin{figure}[htbp]
	\begin{center}
		\includegraphics[scale=0.53]{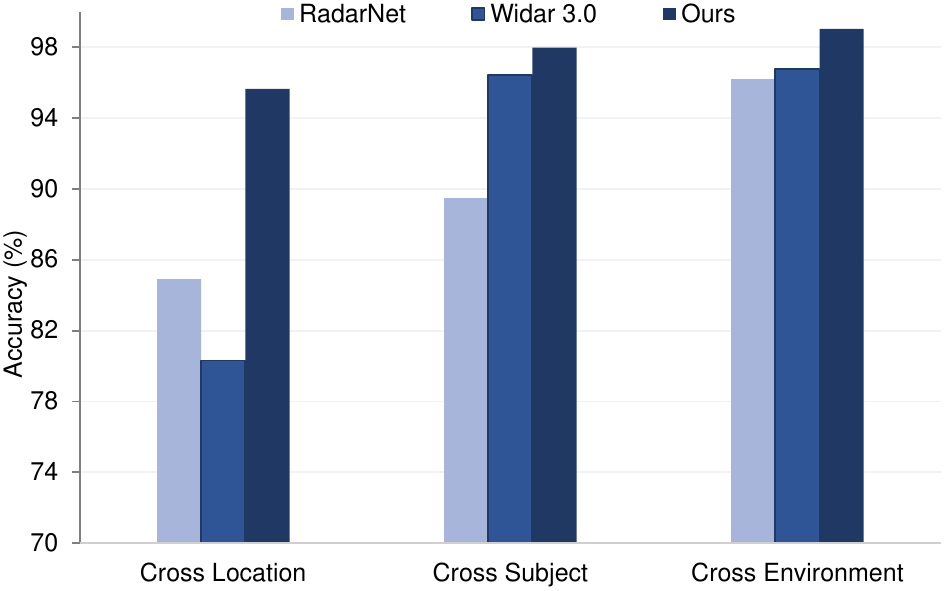}
		\caption{The overall performance of our framework on mmWave dataset.}
		\label{cross domain radar}
	\end{center}
\end{figure}

The overall performance of our framework on mmWave dataset in three domains is shown in Fig. \ref{cross domain radar}. In the figure, the accuracy denotes the average accuracy of every domains, and our framework achieves the highest accuracy in all domains comparing with other baseline methods, which demonstrates the effectiveness of our framework on mmWave dataset. Besides, the detailed performance of our framework in different domains are shown below. The numbers or letters behind the arrow $\rightarrow$ represent the target domain.\par
\emph{(1) Cross Location:} To evaluate our framework's performance of crossing location, we use data of 4 locations as the source domains, and the data of last location as the target domain. In Table \ref{radar location}, L1 - L5 represent the five locations of subject. The results are shown in Table \ref{radar location}. Compared with other methods, our framework achieves the highest average accuracy of 95.65\% and has significant improvement of 15.23\% over the Widar 3.0 and 10.75\% over RadarNet. Location 2 has the highest accuracy of 98.84\%.\par
\begin{table}[htbp]
	\small
	\renewcommand{\arraystretch}{1.5}
	\caption {Cross Location Accuracy (\%)}
	\label{radar location}
	\centering
	\scalebox{0.9}{
		\begin{tabular}{c|c|c|c|c|c|c}
			\hline
			Methods&$\rightarrow$ L1&$\rightarrow$ L2&$\rightarrow$ L3&$\rightarrow$ L4&$\rightarrow$ L5&Avg\\
			\hline
			RadarNet &67.69&96.81&69.77&95.19&95.05&84.90\\
			Widar 3.0 &80.13&97.45&82.59&75.92&65.65&80.32\\
			\hline
			Ours&\textbf{91.20}&\textbf{98.84}&\textbf{94.30}&\textbf{96.94}&\textbf{96.95}&\textbf{95.65}\\
			\hline
	\end{tabular}}
\end{table}
\emph{(2) Cross Subject:} Different subjects' data may show discrepancies due to their unique behavior patterns and body shapes. To evaluate on the subjects, we use data from User 1-6 to evaluate the framework. Each time, we adopt the data from 1 user as the target domain and the other users' data as the source domains. The results in Table \ref{across subject} show that our framework achieves the highest average accuracy of 97.97\%. We also note that all subjects have the accuracy above 93\%, and User 1-2 have the accuracy above 99\%, which demonstrates the impressive performance of our framework.\par
\begin{table}[htbp]
	\small
	\renewcommand{\arraystretch}{1.5}
	\caption {Cross Subject Accuracy (\%)}
	\label{across subject}
	\centering
	\scalebox{0.82}{
		\begin{tabular}{c|c|c|c|c|c|c|c}
			\hline
			Methods&$\rightarrow$U1&$\rightarrow$ U2&$\rightarrow$ U3&$\rightarrow$ U4&$\rightarrow$ U5&$\rightarrow$ U6&Avg\\
			\hline
			RadarNet &95.06&92.93&83.06&85.20&90.26&90.40&89.48\\
			Widar 3.0 &98.80&98.13&91.85&98.33&98.83&92.78&96.45\\
			\hline
			Ours&\textbf{99.42}&\textbf{99.02}&\textbf{93.28}&\textbf{99.33}&98.83&\textbf{96.90}& \textbf{97.97}\\
			\hline
	\end{tabular}}
\end{table}
\emph{(3) Cross Environment:} Different rooms have different layouts, which has an influence on signals propagation. 
To evaluate on the different environments, we use the gesture samples collected in five rooms as the source domains data and the last one as the target domain. E1 - E6 represent the six environments. The results in Table \ref{across environment} show that the framework achieves an overall accuracy of 99.04\%, which is the highest among different methods. 
Moreover, our framework achives an improvement of 2.25\% over the Widar 3.0 and 2.83\% over the RadarNet.\par
\begin{table}[htbp]
	\small
	\renewcommand{\arraystretch}{1.5}
	\caption {Cross Environment Accuracy (\%)}
	\label{across environment}
	\centering
	\scalebox{0.82}{
	\begin{tabular}{c|c|c|c|c|c|c|c}
		\hline
		Methods&$\rightarrow$E1&$\rightarrow$ E2&$\rightarrow$ E3&$\rightarrow$ E4&$\rightarrow$ E5&$\rightarrow$ E6&Avg\\
		\hline
		RadarNet &99.09&97.11&	92.47&	98.66&	98.00&	91.93&	96.21\\
		Widar 3.0 &97.33&98.53&94.80&95.06&98.13&96.93&96.79\\
		\hline
		Ours&\textbf{99.19}&\textbf{98.93}&\textbf{98.66}&\textbf{99.46}&\textbf{98.93}&\textbf{99.06}& \textbf{99.04}\\
		\hline
\end{tabular}}
\end{table}\par
In conclusion, similar to the experiments on WiFi dataset, our framework achieves considerable performance on mmWave radar dataset, which shows that our framework is an effective and general framework that can be used to improve the accuracy of the gesture recognition model with different RF signals.\par 

\subsection{Ablation Experiments}
In this section, we perform extensive ablation studies to evaluate the impact of all modules of the proposed framework.\par
\subsubsection{Impact of Pseudo-labeling and Consistency Regularization}
\begin{figure}[htbp]
	\centering
	\subfigure[The Impact of Pseudo-labeling and Consistency Regularization on WiFi Dataset]{
		\includegraphics[width=0.85 \columnwidth]{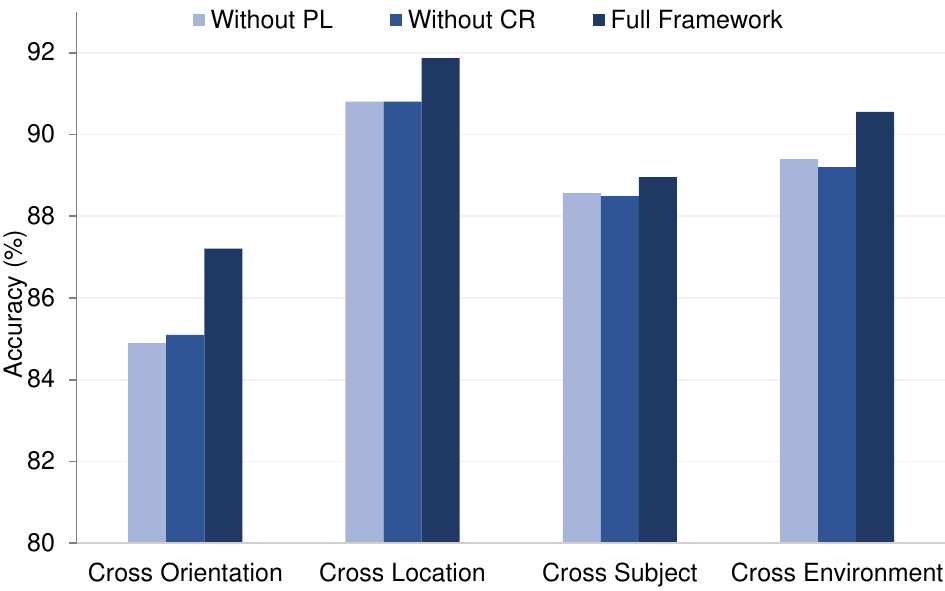}
	}
	\subfigure[The Impact of Pseudo-labeling and Consistency Regularization on mmWave Dataset]{
		\includegraphics[width=0.85 \columnwidth]{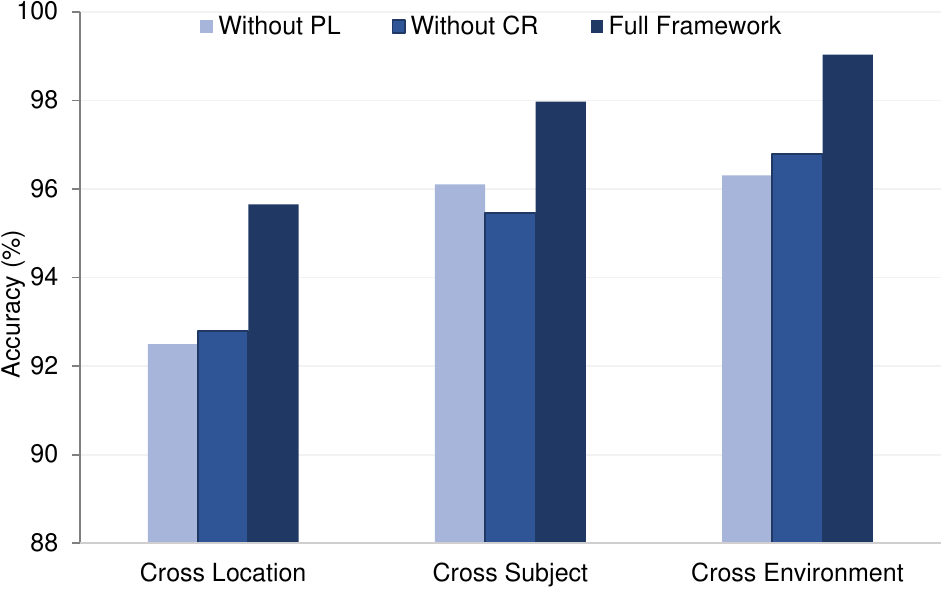}
	}
	\caption{The Impact of Pseudo-labeling and Consistency Regularization.}
	\label{impact of PL CR}
\end{figure}

The performance gain of our framework mainly comes from two modules: pseudo-labeling and consistency regularization. To evaluate which component is essential, we perform two ablation experiments on WiFi dataset and mmWave dataset. Specifically, we remove the consistency regularization module and the pseudo-labeling module from full framework, respectively and keep the other settings unchanged. In Fig. \ref{impact of PL CR}, ``Without CR'' represents the framework whose consistency regularization module is removed, ``Without PL'' represents the framework whose pseudo-labeling module is removed, and the accuracy is the average accuracy of each domain. From Fig. \ref{impact of PL CR}, we can see that the performance of both two incomplete framework is limited, which indicates that both techniques are critical to the proposed framework.\par
\subsubsection{Impact of Cross-match Loss}
\begin{figure}[htbp]
	\centering
	\subfigure[The performance of cross-match loss $L_{u}$ on WiFi dataset.]{
		\includegraphics[width=0.85 \columnwidth]{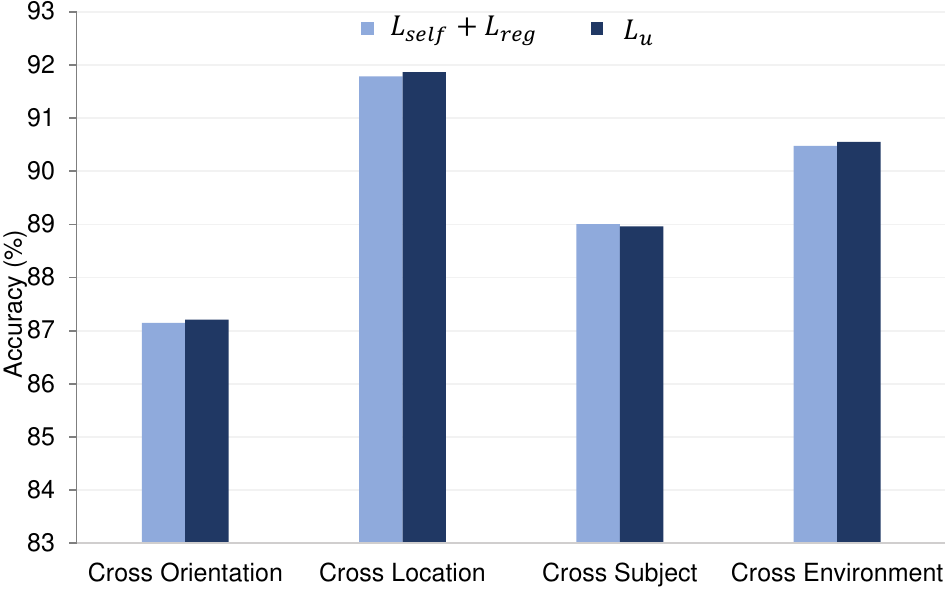}
		\label{cross match wifi editor}
	}
	\subfigure[The performance of cross-match loss $L_{u}$ on mmWave dataset.]{
		\includegraphics[width=0.85 \columnwidth]{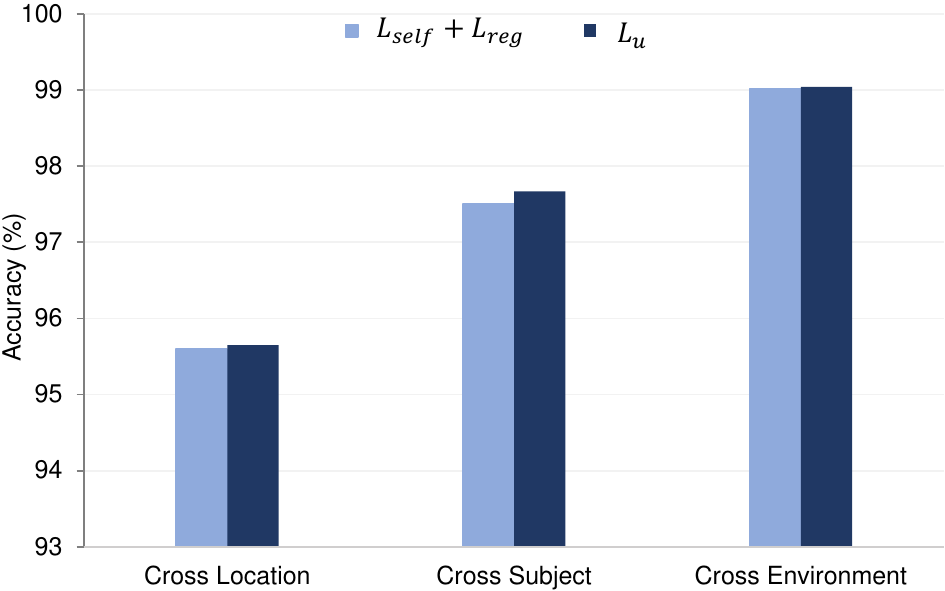}
		\label{cross match radar editor}
	}
	\caption{The performance of cross-match loss $L_{u}$.}
	\label{cross match accuracy editor}
\end{figure}

To better understand the efficiency of the cross-match loss $L_{u}$, we conduct the experiments on WiFi dataset and mmWave dataset on NVIDIA A100-PCIE-40GB GPU. Specifically, we first deploy our framework with the self-supervised loss $L_{self}$ and the consistency regularization loss $L_{reg}$. Then, we deploy our framework only with the cross-match loss $L_{u}$. We keep the other settings unchanged and evaluate the performance in terms of classification accuracy and training time. From Fig .\ref{cross match accuracy editor}, we can see that the classification accuracy of the two methods are so close. Moreover, in Table \ref{time editor}, the framework only with cross-match loss $L_{u}$ needs shorter training time than the framework adopting the two losses $L_{self}$ and $L_{reg}$. In conclusion, the cross-match loss $L_{u}$ makes the whole framework efficient while maintaining the effectiveness.\par

\begin{table}[htbp]
	\small
	\renewcommand{\arraystretch}{1.5}
	\caption {The Training Time of the Two Losses and the Cross-match Loss.}
	\vspace{0.5em}
	\label{time editor}
	\centering
	\scalebox{1}{
		\begin{tabular}{c|c}
			\hline
			Loss&Training Time (second / epoch) \\
			\hline
			$L_{self}$ + $L_{reg}$ &32.5\\
			\hline
			$L_{u}$&\textbf{27.4}\\
			\hline
	\end{tabular}}
\end{table}

\subsubsection{Impact of Confidence Constraint Loss}
\begin{figure}[htbp]
	\centering
	\subfigure[Impact of confidence constraint on WiFi dataset]{
		\includegraphics[width=0.85 \columnwidth]{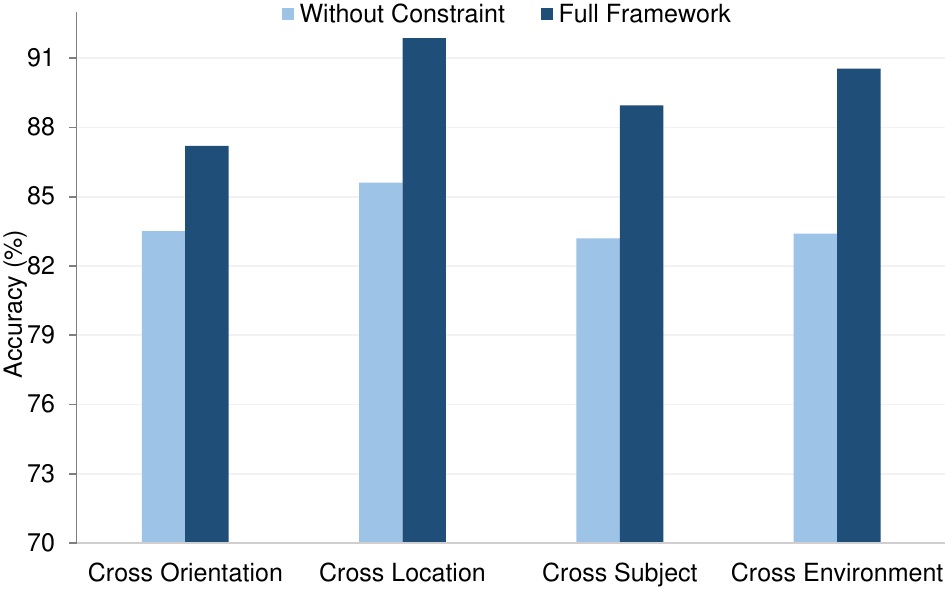}
		\label{impact of constraint WiFi}
	}
	\subfigure[Impact of confidence constraint on mmWave dataset]{
		\includegraphics[width=0.85 \columnwidth]{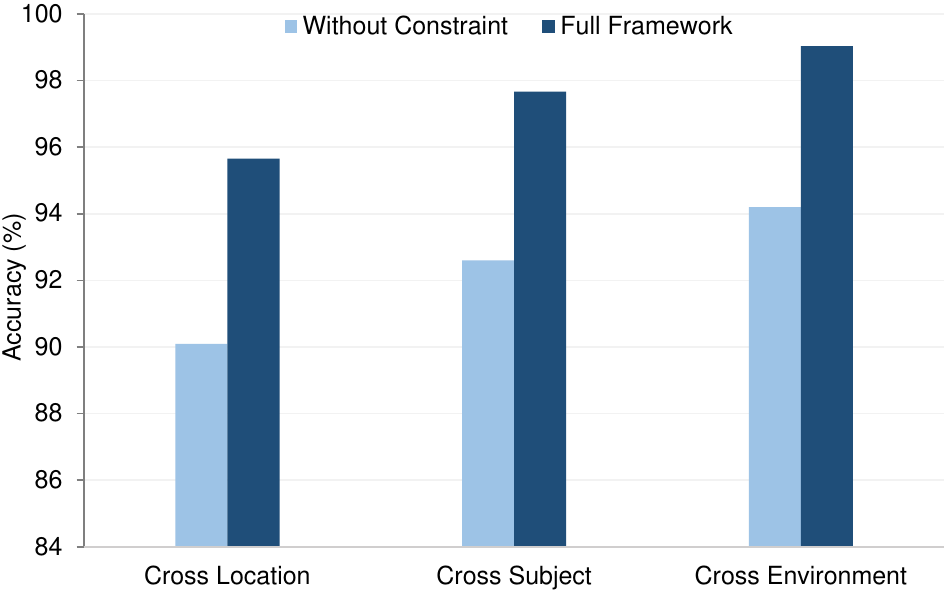}
		\label{impact of constraint radar}
	}
	\caption{Impact of confidence constraint $L_{c}$.}
	\label{impact of constraint}
\end{figure}

In this sub-section, we conduct the ablation experiments on WiFi dataset and mmWave dataset to understand whether our framework can achieve the effective performance without confidence constraint $L_{c}$. To be specific, we compare the full framework with the framework without confidence constraint $L_{c}$. Fig. \ref{impact of constraint} shows that the performance of the framework without the confidence constraint drops rapidly, which indicates that the confidence constraint is necessary and effective to our framework.\par

\subsubsection{Impact of Data Augmentation}
\begin{figure}[htbp]
	\centering
	\subfigure[The accuracy of the WiFi-based methods and our methods.]{
		\includegraphics[width=0.85 \columnwidth]{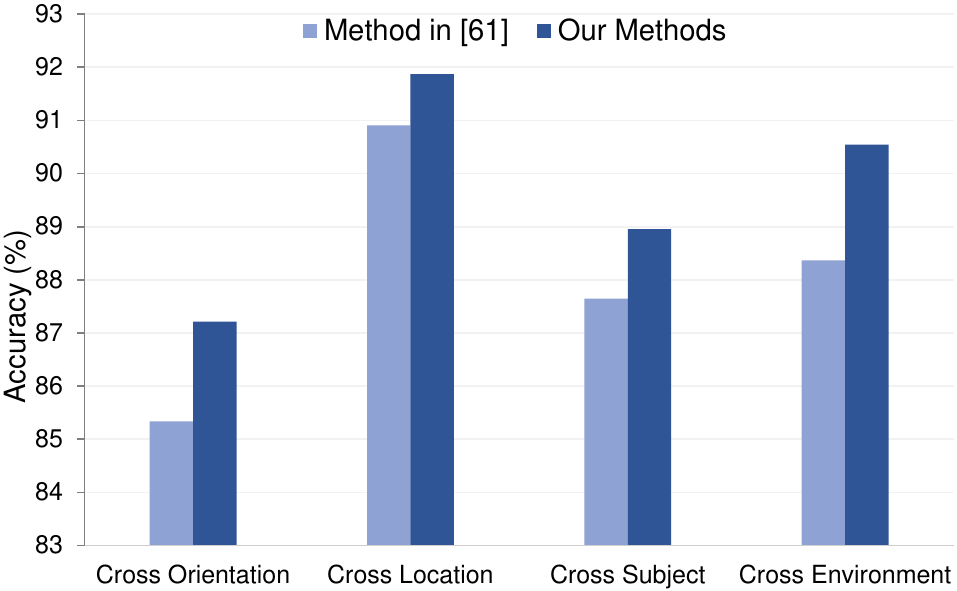}
		\label{data augmentation wifi}
	}
	\subfigure[The accuracy of DI-Gesture-Lite and our methods.]{
		\includegraphics[width=0.85 \columnwidth]{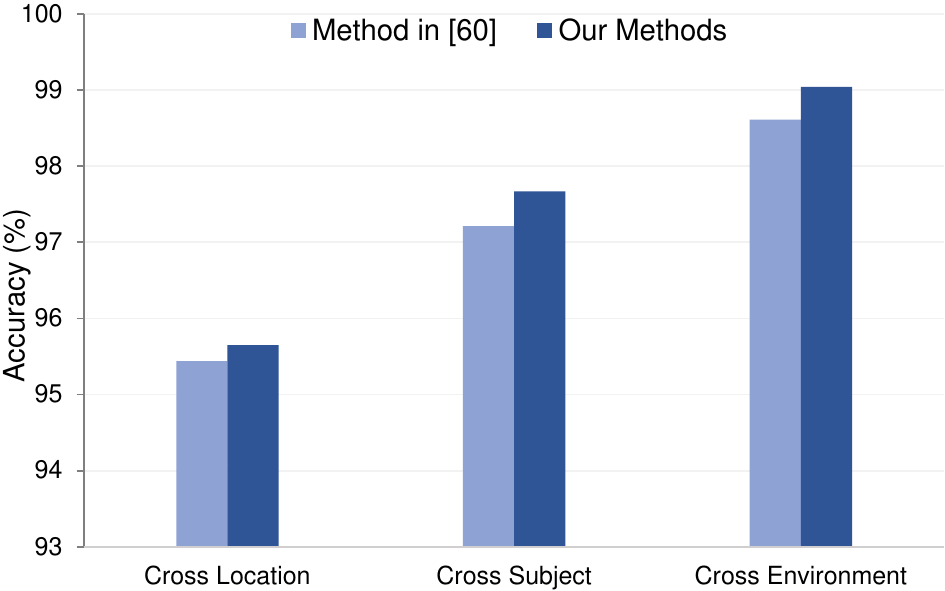}
		\label{data augmentation radar}
	}
	\caption{The comparison with other data augmentation methods.}
	\label{other data augmentation}
\end{figure}

In this work, data augmentation methods are utilized to generate different versions of unlabeled target domain data, which is important to consistency regularization. To demonstrate the effectiveness of the proposed data augmentation method, we compare with WiFi-based method \cite{zhang2020data} and mmWave-based method \cite{li2022towards} on WiFi dataset \cite{zheng2019zero} and mmWave dataset \cite{li2022towards}, and keep other modules of our framework unchanged.
As shown in Fig. \ref{other data augmentation}, the proposed framework with the proposed methods achieves higher accuracy. At the same time, we clearly point out that our augmentation methods are designed for our UDA framework, so that our augmentation methods maybe not applicable to other tasks and other features.\par
\begin{figure}[htbp]
	\centering
	\subfigure[The accuracy of different $N_{l}$ and $N_{t}$ on WiFi dataset.]{
		\includegraphics[width=0.43 \columnwidth]{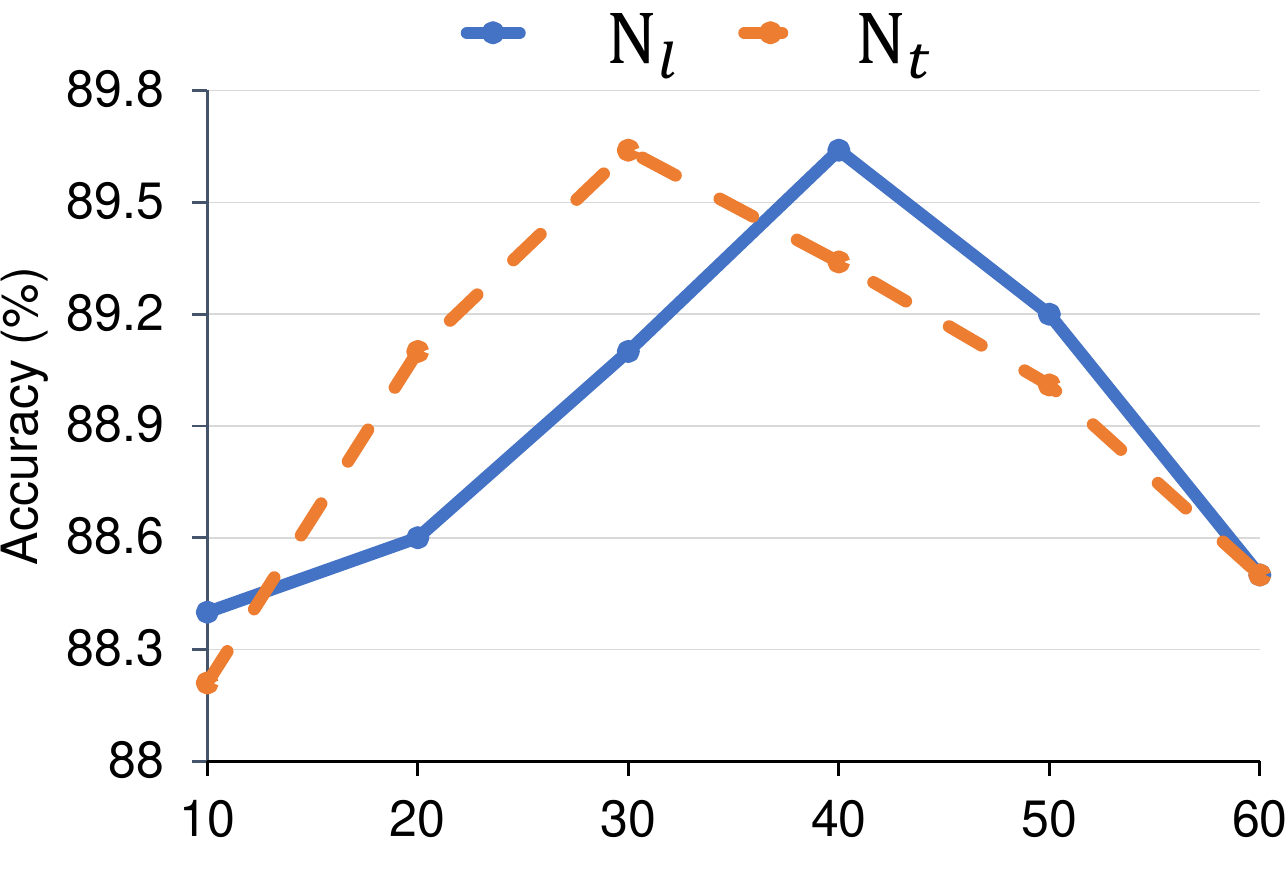 }
		\label{wifi Nl}
	}
	\subfigure[The accuracy of different $N_{l}$ and $N_{t}$ on mmWave dataset.]{
		\includegraphics[width=0.43 \columnwidth]{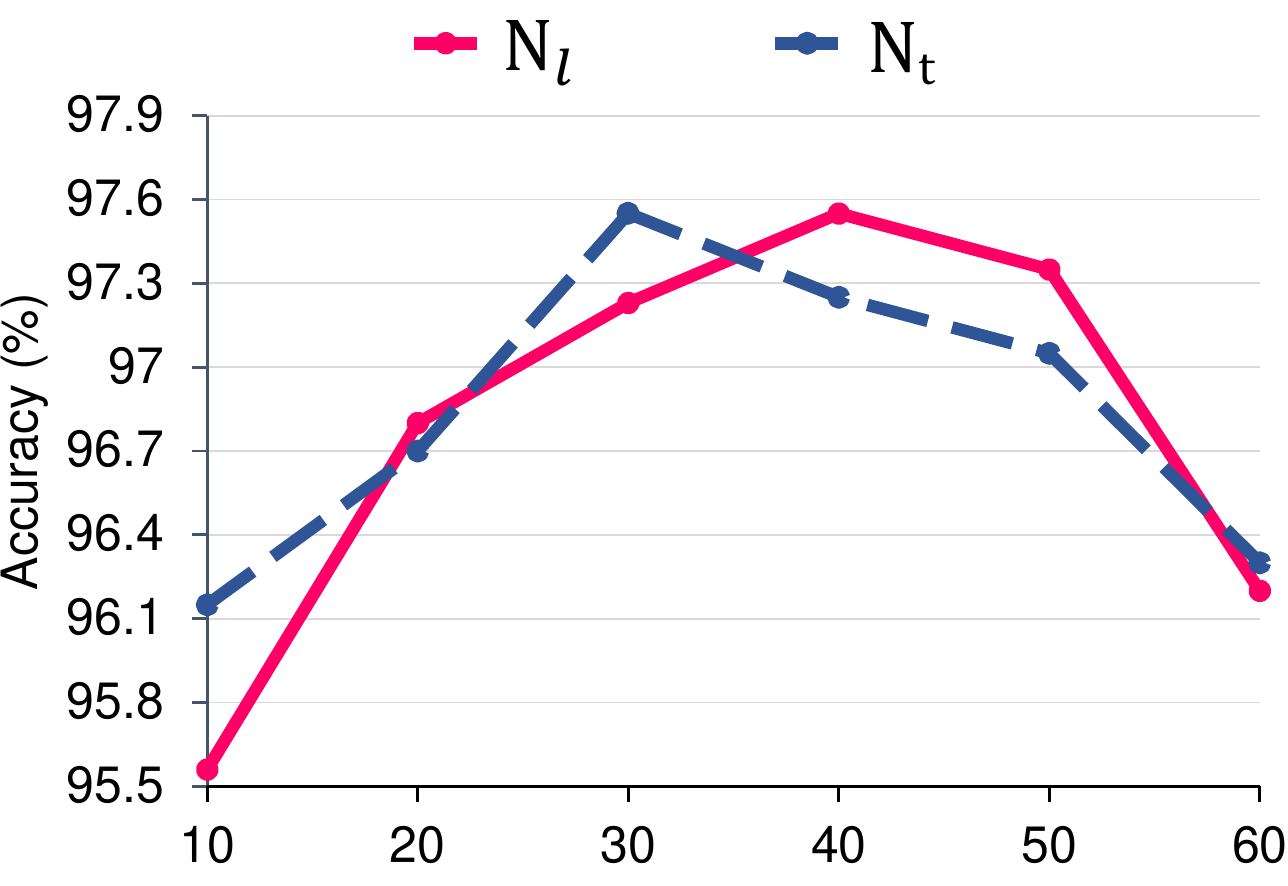}
		\label{wifi Nt}
	}
	\caption{Impact of the number of deleted features.}
	\label{deleted features}
\end{figure}

Furthermore, for local feature erasing, we select $N_{l}$\% informative features randomly and erase them by making their value equal to zero. For time erasing, we remove $N_{t}$\% time frames of the feature matrix randomly. In this work, the $N_{l}$ is experimentally set as 40 and the $N_{t}$ is experimentally set as 30. To investigate the impact of the number of deleted features, we tune $N_{l}$ and $N_{t}$ from 10 to 60 and keep other settings unchanged. 
The Fig. \ref{deleted features} shows that the highest accuracy is obtained when the $N_{l}$ is 40 and the $N_{t}$ is 30. In a nutshell, the deleted features are not the more the better. An appropriate number of the deleted features would enhance the performance of our data augmentation methods.

\subsubsection{Impact of Pseudo Label Threshold}
\begin{figure}[htbp]
	\centering
	\subfigure[The accuracy of different threshold $\tau$ on WiFi dataset.]{
		\includegraphics[width=0.45 \columnwidth]{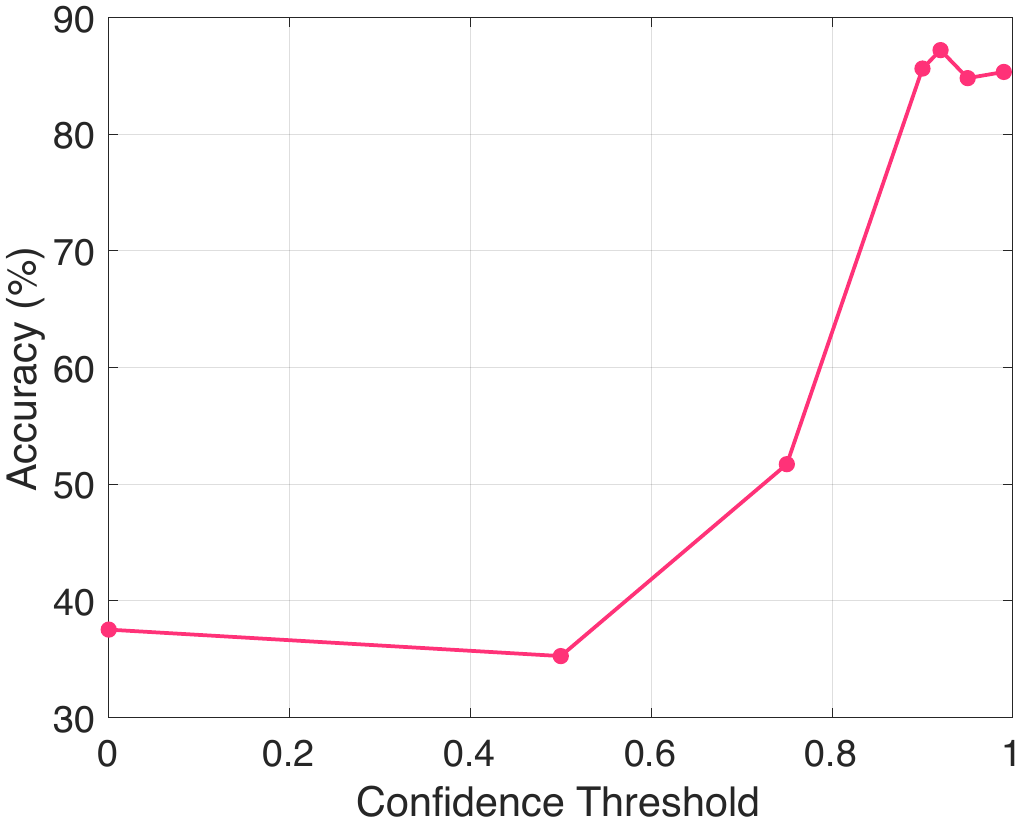}
		\label{Threshold wifi}
	}
	\subfigure[The accuracy of different threshold $\tau$ on mmWave dataset.]{
		\includegraphics[width=0.45 \columnwidth]{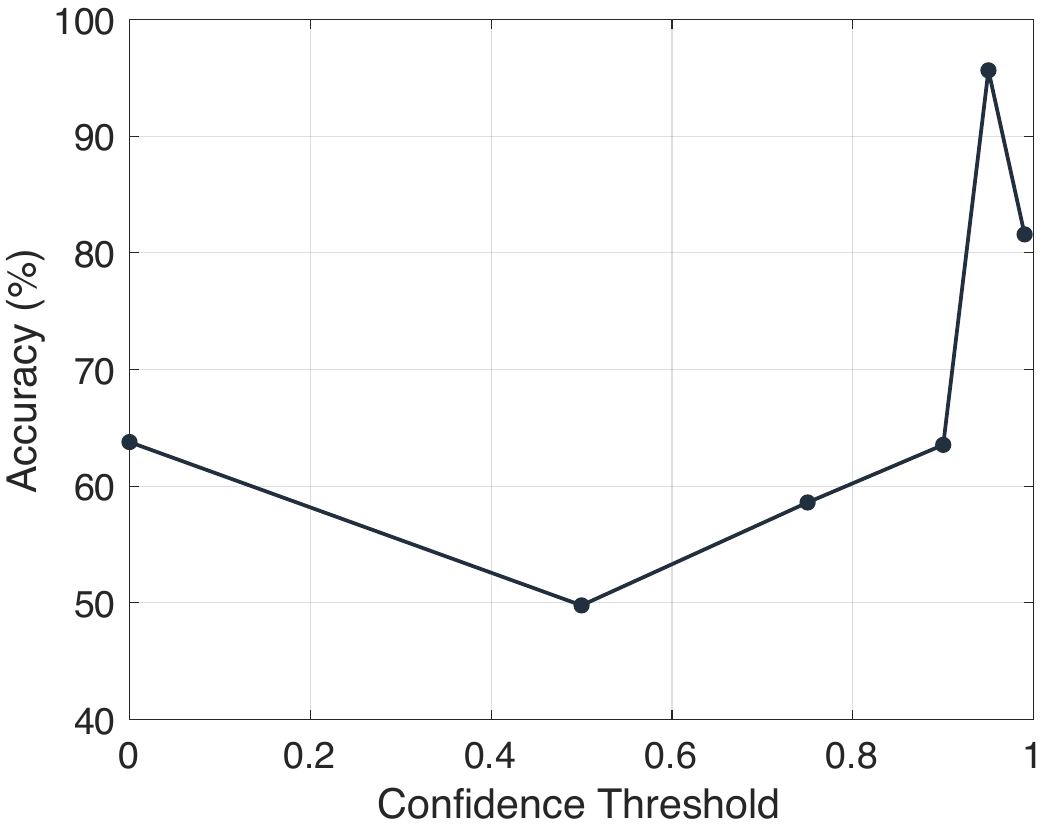}
		\label{Threshold radar}
	}
	\caption{Impact of Confidence Threshold.}
	\label{threshold}
\end{figure}

In our framework, we adopt pseudo-labeling method to obtain the pseudo labels of the unlabeled target domain data. To be specific, we utilize the threshold $\tau$ to select the highly confident predictions as the pseudo labels. In this sub-section, we study the impact of different threshold values. We conduct the analysis experiments on WiFi dataset and mmWave dataset. We set the threshold values ranging from 0 to 0.99 and remain the other settings unchanged. In Fig. \ref{threshold}, the accuracy corresponding to each threshold is the average result of each dataset. The Fig. \ref{Threshold wifi} shows The performance of the framework on WiFi dataset improves with the threshold increasing and achieves the highest accuracy when the threshold value is 0.92. The Fig. \ref{Threshold radar} shows that the highest accuracy on mmWave dataset is obtained when the threshold value is 0.95. In a word, a appropriate threshold is helpful to our framework, which could be obtained experimentally.

\subsection{The efficiency of the proposed framework}
In this section, we investigate the efficiency of the proposed framework. To be specific, 
we compare the Floating Point Operations (FLOPs), training time and testing time of our framework with the backbone model, which is summarized in Table \ref{FLOPs}.
Compared with the backbone, the proposed framework only increase 18.5\% FLOPs. Moreover, we also evaluate the training time per epoch and testing time on NVIDIA A100-PCIE-40GB GPU as shown in Table \ref{FLOPs}. 
From Table. \ref{FLOPs}, we can see that the training time of our framework is slightly longer than that of the backbone, while the testing time keeps the same. We note that there are two reasons about this. One reason is that our method adds the unlabeled target domain data to the training dataset, which leads to more time to train the neural network. The other reason is that the data augmentation operation adds training time.

\begin{table}[htbp]
	\small
	\renewcommand{\arraystretch}{1.5}
	\caption {The FLOPs, training time and testing time of the backbone and our framework.}
	\vspace{0.5em}
	\label{FLOPs}
	\centering
	\scalebox{0.82}{
		\begin{tabular}{c|c|c|c}
			\hline
			Method&FLOPs (M)&Training time (s / epoch)&Testing time (s)\\
			\hline
			Backbone&\textbf{1.03}&\textbf{23.1}&1.01\\
			\hline
			Our framework&1.22&27.4&1.01\\
			\hline
	\end{tabular}}
\end{table}

\section{Conclusion}
\label{conclusion}
In this paper, we proposed an unsupervised domain adaptation framework for RF-based gesture recognition to enhance the performance of the recognition model on the unlabeled target domain. We proposed a pseudo-labeling method and consistency regularization to utilize unlabeled data for model training and align the output distribution for enhancing the robustness of neural network. 
Then we proposed a cross-match loss to integrate the pseudo-labeling and consistency regularization, which makes the whole framework simple yet effective. Moreover, we design a confidence constraint loss and two data augmentation methods to improve the performance of our framework. Finally, we conducted experiments on two different RF gesture recognition testbeds, i.e., WiFi and mmWave signals, to evaluate the performance of the proposed framework. The experimental results demonstrated the superiority of our framework over existing ones. 
We believe that the proposed framework can not only improve the performance of gesture recognition, but also facilitate more investigations towards unsupervised learning with wireless signals.


\begin{thebibliography}{10}
\providecommand{\url}[1]{#1}
\csname url@samestyle\endcsname
\providecommand{\newblock}{\relax}
\providecommand{\bibinfo}[2]{#2}
\providecommand{\BIBentrySTDinterwordspacing}{\spaceskip=0pt\relax}
\providecommand{\BIBentryALTinterwordstretchfactor}{4}
\providecommand{\BIBentryALTinterwordspacing}{\spaceskip=\fontdimen2\font plus
\BIBentryALTinterwordstretchfactor\fontdimen3\font minus
  \fontdimen4\font\relax}
\providecommand{\BIBforeignlanguage}[2]{{%
\expandafter\ifx\csname l@#1\endcsname\relax
\typeout{** WARNING: IEEEtran.bst: No hyphenation pattern has been}%
\typeout{** loaded for the language `#1'. Using the pattern for}%
\typeout{** the default language instead.}%
\else
\language=\csname l@#1\endcsname
\fi
#2}}
\providecommand{\BIBdecl}{\relax}
\BIBdecl

\bibitem{wang2017recurrent}
M.~Wang, B.~Ni, and X.~Yang, ``Recurrent modeling of interaction context for
  collective activity recognition,'' in \emph{Proceedings of the IEEE
  Conference on Computer Vision and Pattern Recognition}, 2017, pp. 3048--3056.

\bibitem{gkioxari2018detecting}
G.~Gkioxari, R.~Girshick, P.~Doll{\'a}r, and K.~He, ``Detecting and recognizing
  human-object interactions,'' in \emph{Proceedings of the IEEE Conference on
  Computer Vision and Pattern Recognition}, 2018, pp. 8359--8367.

\bibitem{li2016practical}
T.~Li, Q.~Liu, and X.~Zhou, ``Practical human sensing in the light,'' in
  \emph{Proceedings of the 14th Annual International Conference on Mobile
  Systems, Applications, and Services}, 2016, pp. 71--84.

\bibitem{bulling2014tutorial}
A.~Bulling, U.~Blanke, and B.~Schiele, ``A tutorial on human activity
  recognition using body-worn inertial sensors,'' \emph{ACM Computing Surveys
  (CSUR)}, vol.~46, no.~3, pp. 1--33, 2014.

\bibitem{guan2017ensembles}
Y.~Guan and T.~Pl{\"o}tz, ``Ensembles of deep lstm learners for activity
  recognition using wearables,'' \emph{Proceedings of the ACM on Interactive,
  Mobile, Wearable and Ubiquitous Technologies}, vol.~1, no.~2, pp. 1--28,
  2017.

\bibitem{shen2016smartwatch}
S.~Shen, H.~Wang, and R.~Roy~Choudhury, ``I am a smartwatch and i can track my
  user's arm,'' in \emph{Proceedings of the 14th annual international
  conference on Mobile systems, applications, and services}, 2016, pp. 85--96.

\bibitem{nandakumar2017covertband}
R.~Nandakumar, A.~Takakuwa, T.~Kohno, and S.~Gollakota, ``Covertband: Activity
  information leakage using music,'' \emph{Proceedings of the ACM on
  Interactive, Mobile, Wearable and Ubiquitous Technologies}, vol.~1, no.~3,
  pp. 1--24, 2017.

\bibitem{yatani2012bodyscope}
K.~Yatani and K.~N. Truong, ``Bodyscope: a wearable acoustic sensor for
  activity recognition,'' in \emph{Proceedings of the 2012 ACM Conference on
  Ubiquitous Computing}, 2012, pp. 341--350.

\bibitem{wang2018device}
J.~Wang, Q.~Gao, M.~Pan, and Y.~Fang, ``Device-free wireless sensing:
  Challenges, opportunities, and applications,'' \emph{IEEE Network}, vol.~32,
  no.~2, pp. 132--137, 2018.

\bibitem{savazzi2016device}
S.~Savazzi, S.~Sigg, M.~Nicoli, V.~Rampa, S.~Kianoush, and U.~Spagnolini,
  ``Device-free radio vision for assisted living: Leveraging wireless channel
  quality information for human sensing,'' \emph{IEEE Signal Processing
  Magazine}, vol.~33, no.~2, pp. 45--58, 2016.

\bibitem{zhou2015sensorless}
Z.~Zhou, C.~Wu, Z.~Yang, and Y.~Liu, ``Sensorless sensing with wifi,''
  \emph{Tsinghua Science and Technology}, vol.~20, no.~1, pp. 1--6, 2015.

\bibitem{wang2014eyes}
Y.~Wang, J.~Liu, Y.~Chen, M.~Gruteser, J.~Yang, and H.~Liu, ``E-eyes:
  device-free location-oriented activity identification using fine-grained wifi
  signatures,'' in \emph{Proceedings of the 20th annual international
  conference on Mobile computing and networking}, 2014, pp. 617--628.

\bibitem{shi2014monitoring}
S.~Shi, S.~Sigg, W.~Zhao, and Y.~Ji, ``Monitoring attention using ambient fm
  radio signals,'' \emph{IEEE Pervasive Computing}, vol.~13, no.~1, pp. 30--36,
  2014.

\bibitem{wang2017device}
W.~Wang, A.~X. Liu, M.~Shahzad, K.~Ling, and S.~Lu, ``Device-free human
  activity recognition using commercial wifi devices,'' \emph{IEEE Journal on
  Selected Areas in Communications}, vol.~35, no.~5, pp. 1118--1131, 2017.

\bibitem{ma2018signfi}
Y.~Ma, G.~Zhou, S.~Wang, H.~Zhao, and W.~Jung, ``Signfi: Sign language
  recognition using wifi,'' \emph{Proceedings of the ACM on Interactive,
  Mobile, Wearable and Ubiquitous Technologies}, vol.~2, no.~1, pp. 1--21,
  2018.

\bibitem{virmani2017position}
A.~Virmani and M.~Shahzad, ``Position and orientation agnostic gesture
  recognition using wifi,'' in \emph{Proceedings of the 15th Annual
  International Conference on Mobile Systems, Applications, and Services},
  2017, pp. 252--264.

\bibitem{zheng2019zero}
Y.~Zheng, Y.~Zhang, K.~Qian, G.~Zhang, Y.~Liu, C.~Wu, and Z.~Yang,
  ``Zero-effort cross-domain gesture recognition with wi-fi,'' in
  \emph{Proceedings of the 17th Annual International Conference on Mobile
  Systems, Applications, and Services}, 2019, pp. 313--325.

\bibitem{zhang2018crosssense}
J.~Zhang, Z.~Tang, M.~Li, D.~Fang, P.~Nurmi, and Z.~Wang, ``Crosssense: Towards
  cross-site and large-scale wifi sensing,'' in \emph{Proceedings of the 24th
  Annual International Conference on Mobile Computing and Networking}, 2018,
  pp. 305--320.

\bibitem{jiang2018towards}
W.~Jiang, C.~Miao, F.~Ma, S.~Yao, Y.~Wang, Y.~Yuan, H.~Xue, C.~Song, X.~Ma,
  D.~Koutsonikolas \emph{et~al.}, ``Towards environment independent device free
  human activity recognition,'' in \emph{Proceedings of the 24th Annual
  International Conference on Mobile Computing and Networking}, 2018, pp.
  289--304.

\bibitem{xiao2019csigan}
C.~Xiao, D.~Han, Y.~Ma, and Z.~Qin, ``Csigan: Robust channel state
  information-based activity recognition with gans,'' \emph{IEEE Internet of
  Things Journal}, vol.~6, no.~6, pp. 10\,191--10\,204, 2019.

\bibitem{wang2019wicar}
F.~Wang, J.~Liu, and W.~Gong, ``Wicar: Wifi-based in-car activity recognition
  with multi-adversarial domain adaptation,'' in \emph{Proceedings of the
  International Symposium on Quality of Service}, 2019, pp. 1--10.

\bibitem{wang2021environment}
Z.~Wang, S.~Chen, W.~Yang, and Y.~Xu, ``Environment-independent wi-fi human
  activity recognition with adversarial network,'' in \emph{ICASSP 2021-2021
  IEEE International Conference on Acoustics, Speech and Signal Processing
  (ICASSP)}.\hskip 1em plus 0.5em minus 0.4em\relax IEEE, 2021, pp. 3330--3334.

\bibitem{kang2021context}
H.~Kang, Q.~Zhang, and Q.~Huang, ``Context-aware wireless based cross domain
  gesture recognition,'' \emph{IEEE Internet of Things Journal}, 2021.

\bibitem{ma2019practical}
X.~Ma, Y.~Zhao, L.~Zhang, Q.~Gao, M.~Pan, and J.~Wang, ``Practical device-free
  gesture recognition using wifi signals based on metalearning,'' \emph{IEEE
  Transactions on Industrial Informatics}, vol.~16, no.~1, pp. 228--237, 2019.

\bibitem{sajjadi2016regularization}
M.~Sajjadi, M.~Javanmardi, and T.~Tasdizen, ``Regularization with stochastic
  transformations and perturbations for deep semi-supervised learning,''
  \emph{Advances in neural information processing systems}, vol.~29, pp.
  1163--1171, 2016.

\bibitem{tarvainen2017mean}
A.~Tarvainen and H.~Valpola, ``Mean teachers are better role models:
  Weight-averaged consistency targets improve semi-supervised deep learning
  results,'' \emph{arXiv preprint arXiv:1703.01780}, 2017.

\bibitem{sohn2020fixmatch}
K.~Sohn, D.~Berthelot, C.-L. Li, Z.~Zhang, N.~Carlini, E.~D. Cubuk, A.~Kurakin,
  H.~Zhang, and C.~Raffel, ``Fixmatch: Simplifying semi-supervised learning
  with consistency and confidence,'' \emph{arXiv preprint arXiv:2001.07685},
  2020.

\bibitem{xie2019unsupervised}
Q.~Xie, Z.~Dai, E.~Hovy, M.-T. Luong, and Q.~V. Le, ``Unsupervised data
  augmentation for consistency training,'' \emph{arXiv preprint
  arXiv:1904.12848}, 2019.

\bibitem{lee2013pseudo}
D.-H. Lee \emph{et~al.}, ``Pseudo-label: The simple and efficient
  semi-supervised learning method for deep neural networks,'' in \emph{Workshop
  on challenges in representation learning, ICML}, vol.~3, no.~2, 2013, p. 896.

\bibitem{gopalan2011domain}
R.~Gopalan, R.~Li, and R.~Chellappa, ``Domain adaptation for object
  recognition: An unsupervised approach,'' in \emph{2011 international
  conference on computer vision}.\hskip 1em plus 0.5em minus 0.4em\relax IEEE,
  2011, pp. 999--1006.

\bibitem{na2021fixbi}
J.~Na, H.~Jung, H.~J. Chang, and W.~Hwang, ``Fixbi: Bridging domain spaces for
  unsupervised domain adaptation,'' in \emph{Proceedings of the IEEE/CVF
  Conference on Computer Vision and Pattern Recognition}, 2021, pp. 1094--1103.

\bibitem{long2016unsupervised}
M.~Long, H.~Zhu, J.~Wang, and M.~I. Jordan, ``Unsupervised domain adaptation
  with residual transfer networks,'' \emph{arXiv preprint arXiv:1602.04433},
  2016.

\bibitem{shorten2019survey}
C.~Shorten and T.~M. Khoshgoftaar, ``A survey on image data augmentation for
  deep learning,'' \emph{Journal of Big Data}, vol.~6, no.~1, pp. 1--48, 2019.

\bibitem{zhang2020data}
J.~Zhang, F.~Wu, B.~Wei, Q.~Zhang, H.~Huang, S.~W. Shah, and J.~Cheng, ``Data
  augmentation and dense-lstm for human activity recognition using wifi
  signal,'' \emph{IEEE Internet of Things Journal}, vol.~8, no.~6, pp.
  4628--4641, 2020.

\bibitem{qian1999momentum}
N.~Qian, ``On the momentum term in gradient descent learning algorithms,''
  \emph{Neural networks}, vol.~12, no.~1, pp. 145--151, 1999.

\bibitem{wang2016lifs}
J.~Wang, H.~Jiang, J.~Xiong, K.~Jamieson, X.~Chen, D.~Fang, and B.~Xie, ``Lifs:
  Low human-effort, device-free localization with fine-grained subcarrier
  information,'' in \emph{Proceedings of the 22nd Annual International
  Conference on Mobile Computing and Networking}, 2016, pp. 243--256.

\bibitem{li2017indotrack}
X.~Li, D.~Zhang, Q.~Lv, J.~Xiong, S.~Li, Y.~Zhang, and H.~Mei, ``Indotrack:
  Device-free indoor human tracking with commodity wi-fi,'' \emph{Proceedings
  of the ACM on Interactive, Mobile, Wearable and Ubiquitous Technologies},
  vol.~1, no.~3, pp. 1--22, 2017.

\bibitem{kalgaonkar2009one}
K.~Kalgaonkar and B.~Raj, ``One-handed gesture recognition using ultrasonic
  doppler sonar,'' in \emph{2009 IEEE International Conference on Acoustics,
  Speech and Signal Processing}.\hskip 1em plus 0.5em minus 0.4em\relax IEEE,
  2009, pp. 1889--1892.

\bibitem{huang2016indoor}
X.~Huang and M.~Dai, ``Indoor device-free activity recognition based on radio
  signal,'' \emph{IEEE Transactions on Vehicular Technology}, vol.~66, no.~6,
  pp. 5316--5329, 2016.

\bibitem{lien2016soli}
J.~Lien, N.~Gillian, M.~E. Karagozler, P.~Amihood, C.~Schwesig, E.~Olson,
  H.~Raja, and I.~Poupyrev, ``Soli: Ubiquitous gesture sensing with millimeter
  wave radar,'' \emph{ACM Transactions on Graphics (TOG)}, vol.~35, no.~4, pp.
  1--19, 2016.

\bibitem{zhu2018indoor}
S.~Zhu, J.~Xu, H.~Guo, Q.~Liu, S.~Wu, and H.~Wang, ``Indoor human activity
  recognition based on ambient radar with signal processing and machine
  learning,'' in \emph{2018 IEEE international conference on communications
  (ICC)}.\hskip 1em plus 0.5em minus 0.4em\relax IEEE, 2018, pp. 1--6.

\bibitem{fan2016wireless}
T.~Fan, C.~Ma, Z.~Gu, Q.~Lv, J.~Chen, D.~Ye, J.~Huangfu, Y.~Sun, C.~Li, and
  L.~Ran, ``Wireless hand gesture recognition based on continuous-wave doppler
  radar sensors,'' \emph{IEEE Transactions on Microwave Theory and Techniques},
  vol.~64, no.~11, pp. 4012--4020, 2016.

\bibitem{zhu2005semi}
X.~J. Zhu, ``Semi-supervised learning literature survey,'' 2005.

\bibitem{liu2020real}
H.~Liu, Y.~Wang, A.~Zhou, H.~He, W.~Wang, K.~Wang, P.~Pan, Y.~Lu, L.~Liu, and
  H.~Ma, ``Real-time arm gesture recognition in smart home scenarios via
  millimeter wave sensing,'' \emph{Proceedings of the ACM on Interactive,
  Mobile, Wearable and Ubiquitous Technologies}, vol.~4, no.~4, pp. 1--28,
  2020.

\bibitem{santhalingam2020mmasl}
P.~S. Santhalingam, A.~A. Hosain, D.~Zhang, P.~Pathak, H.~Rangwala, and
  R.~Kushalnagar, ``mmasl: Environment-independent asl gesture recognition
  using 60 ghz millimeter-wave signals,'' \emph{Proceedings of the ACM on
  Interactive, Mobile, Wearable and Ubiquitous Technologies}, vol.~4, no.~1,
  pp. 1--30, 2020.

\bibitem{hayashi2021radarnet}
E.~Hayashi, J.~Lien, N.~Gillian, L.~Giusti, D.~Weber, J.~Yamanaka, L.~Bedal,
  and I.~Poupyrev, ``Radarnet: Efficient gesture recognition technique
  utilizing a miniature radar sensor,'' in \emph{Proceedings of the 2021 CHI
  Conference on Human Factors in Computing Systems}, 2021, pp. 1--14.

\bibitem{berthelot2019remixmatch}
D.~Berthelot, N.~Carlini, E.~D. Cubuk, A.~Kurakin, K.~Sohn, H.~Zhang, and
  C.~Raffel, ``Remixmatch: Semi-supervised learning with distribution alignment
  and augmentation anchoring,'' \emph{arXiv preprint arXiv:1911.09785}, 2019.

\bibitem{berthelot2019mixmatch}
D.~Berthelot, N.~Carlini, I.~Goodfellow, N.~Papernot, A.~Oliver, and C.~Raffel,
  ``Mixmatch: A holistic approach to semi-supervised learning,'' \emph{arXiv
  preprint arXiv:1905.02249}, 2019.

\bibitem{zhai2019s4l}
X.~Zhai, A.~Oliver, A.~Kolesnikov, and L.~Beyer, ``S4l: Self-supervised
  semi-supervised learning,'' in \emph{Proceedings of the IEEE/CVF
  International Conference on Computer Vision}, 2019, pp. 1476--1485.

\bibitem{ao2017fast}
S.~Ao, X.~Li, and C.~Ling, ``Fast generalized distillation for semi-supervised
  domain adaptation,'' in \emph{Proceedings of the AAAI Conference on
  Artificial Intelligence}, vol.~31, no.~1, 2017.

\bibitem{saito2019semi}
K.~Saito, D.~Kim, S.~Sclaroff, T.~Darrell, and K.~Saenko, ``Semi-supervised
  domain adaptation via minimax entropy,'' in \emph{Proceedings of the IEEE/CVF
  International Conference on Computer Vision}, 2019, pp. 8050--8058.

\bibitem{qin2021contradictory}
C.~Qin, L.~Wang, Q.~Ma, Y.~Yin, H.~Wang, and Y.~Fu, ``Contradictory structure
  learning for semi-supervised domain adaptation,'' in \emph{Proceedings of the
  2021 SIAM International Conference on Data Mining (SDM)}.\hskip 1em plus
  0.5em minus 0.4em\relax SIAM, 2021, pp. 576--584.

\bibitem{li2020online}
D.~Li and T.~Hospedales, ``Online meta-learning for multi-source and
  semi-supervised domain adaptation,'' in \emph{European Conference on Computer
  Vision}.\hskip 1em plus 0.5em minus 0.4em\relax Springer, 2020, pp. 382--403.

\bibitem{yang2020mico}
L.~Yang, Y.~Wang, M.~Gao, A.~Shrivastava, K.~Q. Weinberger, W.-L. Chao, and
  S.-N. Lim, ``Mico: Mixup co-training for semi-supervised domain adaptation,''
  \emph{arXiv e-prints}, pp. arXiv--2007, 2020.

\bibitem{ben2007analysis}
S.~Ben-David, J.~Blitzer, K.~Crammer, F.~Pereira \emph{et~al.}, ``Analysis of
  representations for domain adaptation,'' \emph{Advances in neural information
  processing systems}, vol.~19, p. 137, 2007.

\bibitem{jhuo2012robust}
I.-H. Jhuo, D.~Liu, D.~Lee, and S.-F. Chang, ``Robust visual domain adaptation
  with low-rank reconstruction,'' in \emph{2012 IEEE conference on computer
  vision and pattern recognition}.\hskip 1em plus 0.5em minus 0.4em\relax IEEE,
  2012, pp. 2168--2175.

\bibitem{long2015learning}
M.~Long, Y.~Cao, J.~Wang, and M.~Jordan, ``Learning transferable features with
  deep adaptation networks,'' in \emph{International conference on machine
  learning}.\hskip 1em plus 0.5em minus 0.4em\relax PMLR, 2015, pp. 97--105.

\bibitem{long2017deep}
M.~Long, H.~Zhu, J.~Wang, and M.~I. Jordan, ``Deep transfer learning with joint
  adaptation networks,'' in \emph{International conference on machine
  learning}.\hskip 1em plus 0.5em minus 0.4em\relax PMLR, 2017, pp. 2208--2217.

\bibitem{kang2019contrastive}
G.~Kang, L.~Jiang, Y.~Yang, and A.~G. Hauptmann, ``Contrastive adaptation
  network for unsupervised domain adaptation,'' in \emph{Proceedings of the
  IEEE/CVF Conference on Computer Vision and Pattern Recognition}, 2019, pp.
  4893--4902.

\bibitem{tzeng2017adversarial}
E.~Tzeng, J.~Hoffman, K.~Saenko, and T.~Darrell, ``Adversarial discriminative
  domain adaptation,'' in \emph{Proceedings of the IEEE conference on computer
  vision and pattern recognition}, 2017, pp. 7167--7176.

\bibitem{sankaranarayanan2018generate}
S.~Sankaranarayanan, Y.~Balaji, C.~D. Castillo, and R.~Chellappa, ``Generate to
  adapt: Aligning domains using generative adversarial networks,'' in
  \emph{Proceedings of the IEEE Conference on Computer Vision and Pattern
  Recognition}, 2018, pp. 8503--8512.

\bibitem{zhu2017unpaired}
J.-Y. Zhu, T.~Park, P.~Isola, and A.~A. Efros, ``Unpaired image-to-image
  translation using cycle-consistent adversarial networks,'' in
  \emph{Proceedings of the IEEE international conference on computer vision},
  2017, pp. 2223--2232.

\bibitem{ganin2016domain}
Y.~Ganin, E.~Ustinova, H.~Ajakan, P.~Germain, H.~Larochelle, F.~Laviolette,
  M.~Marchand, and V.~Lempitsky, ``Domain-adversarial training of neural
  networks,'' \emph{The journal of machine learning research}, vol.~17, no.~1,
  pp. 2096--2030, 2016.

\bibitem{liu2017unsupervised}
M.-Y. Liu, T.~Breuel, and J.~Kautz, ``Unsupervised image-to-image translation
  networks,'' in \emph{Advances in neural information processing systems},
  2017, pp. 700--708.

\bibitem{zhang2020mtrack}
D.~Zhang, Y.~Hu, and Y.~Chen, ``Mtrack: Tracking multiperson moving
  trajectories and vital signs with radio signals,'' \emph{IEEE Internet of
  Things Journal}, vol.~8, no.~5, pp. 3904--3914, 2020.

\bibitem{chen2020speednet}
Y.~Chen, H.~Deng, D.~Zhang, and Y.~Hu, ``Speednet: Indoor speed estimation with
  radio signals,'' \emph{IEEE Internet of Things Journal}, vol.~8, no.~4, pp.
  2762--2774, 2020.

\bibitem{he2020wifi}
Y.~He, Y.~Chen, Y.~Hu, and B.~Zeng, ``Wifi vision: sensing, recognition, and
  detection with commodity mimo-ofdm wifi,'' \emph{IEEE Internet of Things
  Journal}, vol.~7, no.~9, pp. 8296--8317, 2020.

\bibitem{zhang2019breathtrack}
D.~Zhang, Y.~Hu, Y.~Chen, and B.~Zeng, ``Breathtrack: Tracking indoor human
  breath status via commodity wifi,'' \emph{IEEE Internet of Things Journal},
  vol.~6, no.~2, pp. 3899--3911, 2019.

\bibitem{zhang2018multitarget}
D.~Zhang, Y.~He, X.~Gong, Y.~Hu, Y.~Chen, and B.~Zeng, ``Multitarget aoa
  estimation using wideband lfmcw signal and two receiver antennas,''
  \emph{IEEE Transactions on Vehicular Technology}, vol.~67, no.~8, pp.
  7101--7112, 2018.

\bibitem{he2020non}
Y.~He, D.~Zhang, Y.~Hu, and Y.~Chen, ``Non-line-of-sight imaging with radio
  signals,'' in \emph{2020 Asia-Pacific Signal and Information Processing
  Association Annual Summit and Conference (APSIPA ASC)}.\hskip 1em plus 0.5em
  minus 0.4em\relax IEEE, 2020, pp. 11--16.

\bibitem{li2020wihf}
C.~L. Li, M.~Liu, and Z.~Cao, ``Wihf: Gesture and user recognition with wifi,''
  \emph{IEEE Transactions on Mobile Computing}, 2020.

\bibitem{zou2016grfid}
Y.~Zou, J.~Xiao, J.~Han, K.~Wu, Y.~Li, and L.~M. Ni, ``Grfid: A device-free
  rfid-based gesture recognition system,'' \emph{IEEE Transactions on Mobile
  Computing}, vol.~16, no.~2, pp. 381--393, 2016.

\bibitem{abdelnasser2018ubiquitous}
H.~Abdelnasser, K.~Harras, and M.~Youssef, ``A ubiquitous wifi-based
  fine-grained gesture recognition system,'' \emph{IEEE Transactions on Mobile
  Computing}, vol.~18, no.~11, pp. 2474--2487, 2018.

\bibitem{zhang2019calibrating}
D.~Zhang, Y.~Hu, Y.~Chen, and B.~Zeng, ``Calibrating phase offsets for
  commodity wifi,'' \emph{IEEE Systems Journal}, vol.~14, no.~1, pp. 661--664,
  2019.

\bibitem{xu2017trieds}
Q.~Xu, Y.~Chen, B.~Wang, and K.~R. Liu, ``Trieds: Wireless events detection
  through the wall,'' \emph{IEEE Internet of Things Journal}, vol.~4, no.~3,
  pp. 723--735, 2017.

\bibitem{han2016enabling}
Y.~Han, Y.~Chen, B.~Wang, and K.~R. Liu, ``Enabling heterogeneous connectivity
  in internet of things: A time-reversal approach,'' \emph{IEEE Internet of
  Things Journal}, vol.~3, no.~6, pp. 1036--1047, 2016.

\bibitem{xu2017radio}
Q.~Xu, Y.~Chen, B.~Wang, and K.~R. Liu, ``Radio biometrics: Human recognition
  through a wall,'' \emph{IEEE Transactions on Information Forensics and
  Security}, vol.~12, no.~5, pp. 1141--1155, 2017.

\bibitem{chen2019residual}
Y.~Chen, X.~Su, Y.~Hu, and B.~Zeng, ``Residual carrier frequency offset
  estimation and compensation for commodity wifi,'' \emph{IEEE Transactions on
  Mobile Computing}, vol.~19, no.~12, pp. 2891--2902, 2019.

\bibitem{li2021learning}
J.~Li, Z.~Wang, and X.~Hu, ``Learning intact features by erasing-inpainting for
  few-shot classification,'' in \emph{Proceedings of the AAAI Conference on
  Artificial Intelligence}, vol.~35, no.~9, 2021, pp. 8401--8409.

\bibitem{li2022towards}
Y.~Li, D.~Zhang, J.~Chen, J.~Wan, D.~Zhang, Y.~Hu, Q.~Sun, and Y.~Chen,
  ``Towards domain-independent and real-time gesture recognition using mmwave
  signal,'' \emph{IEEE Transactions on Mobile Computing}, 2022.

\bibitem{zou2019confidence}
Y.~Zou, Z.~Yu, X.~Liu, B.~Kumar, and J.~Wang, ``Confidence regularized
  self-training,'' in \emph{Proceedings of the IEEE/CVF International
  Conference on Computer Vision}, 2019, pp. 5982--5991.

\bibitem{arazo2020pseudo}
E.~Arazo, D.~Ortego, P.~Albert, N.~E. O’Connor, and K.~McGuinness,
  ``Pseudo-labeling and confirmation bias in deep semi-supervised learning,''
  in \emph{2020 International Joint Conference on Neural Networks
  (IJCNN)}.\hskip 1em plus 0.5em minus 0.4em\relax IEEE, 2020, pp. 1--8.

\bibitem{bagherinezhad2018label}
H.~Bagherinezhad, M.~Horton, M.~Rastegari, and A.~Farhadi, ``Label refinery:
  Improving imagenet classification through label progression,'' \emph{arXiv
  preprint arXiv:1805.02641}, 2018.

\bibitem{robey1992cfar}
F.~C. Robey, D.~R. Fuhrmann, E.~J. Kelly, and R.~Nitzberg, ``A cfar adaptive
  matched filter detector,'' \emph{IEEE Transactions on aerospace and
  electronic systems}, vol.~28, no.~1, pp. 208--216, 1992.

\end{thebibliography}
\end{document}